\documentclass[10pt,twocolumn,letterpaper]{article}

\usepackage{iccv}
\usepackage{times}
\usepackage{epsfig}
\usepackage{graphicx}
\usepackage{amsmath}
\usepackage{amssymb}
\usepackage{multirow}
\usepackage{colortbl}
\usepackage{color}
\usepackage{threeparttable}
\usepackage{booktabs}
\usepackage{adjustbox}
\usepackage{subfig}
\usepackage{caption}

\usepackage[pagebackref=true,breaklinks=true,letterpaper=true,colorlinks,bookmarks=false]{hyperref}

\iccvfinalcopy 

\def\iccvPaperID{1794} 

\ificcvfinal\fi
\newcommand{\pp}{p}

\newcommand{\RR}{\mathbb{R}}

\begin{document}
	
	\title{BoundarySqueeze: Image Segmentation as Boundary Squeezing}
	
	\author{
		Hao He$^{1,2}$\thanks{Equal Contribution, Email: hehao2019@ia.ac.cn, lxtpku@pku.edu.cn},
		Xiangtai Li$^{3*}$,
		Yibo Yang$^{3}$,
		Guangliang Cheng$^{4}$, \\
		Yunhai Tong$^3$,
		Lubin Weng$^{1,2}$,
		Zhouchen Lin$^{3}$,
		Shiming Xiang$^{1,2}$
		\\[0.2cm]
		\small $ ^1$ National Laboratory of Pattern Recognition, Institute of Automation, Chinese Academy of Sciences \\
		\small $ ^2$ School of Artificial Intelligence, University of Chinese Academy of Sciences \\
		\small $ ^3$ Key Laboratory of Machine Perception (MOE), Peking University
		\small $ ^4$ SenseTime Research
	}

	\maketitle
	\ificcvfinal\thispagestyle{empty}\fi
	
	
	\begin{abstract}
This paper proposes a novel method for high-quality image segmentation of both objects and scenes. Inspired by the dilation and erosion operations in morphological image processing techniques, the pixel-level image segmentation problems are treated as squeezing object boundaries. From this perspective, a novel and efficient \textbf{Boundary Squeeze} module is proposed. This module is used to squeeze the object boundary from both inner and outer directions, which contributes to precise mask representation. A bi-directionally flow-based warping process is proposed to generate such squeezed feature representation, and two specific loss signals are designed to supervise the squeezing process. The Boundary Squeeze module can be easily applied to both instance and semantic segmentation tasks as a plug-and-play module by building on top of some existing methods. Moreover, the proposed module is light-weighted, and thus has potential for practical usage. Experiment results show that our simple yet effective design can produce high-quality results on several different datasets. Besides, several other metrics on the boundary are used to prove the effectiveness of our method over previous work. Our approach yields significant improvement on challenging COCO and Cityscapes datasets for both instance and semantic segmentation, and outperforms previous state-of-the-art PointRend in both accuracy and speed under the same setting. Codes and models will be published at \url{https://github.com/lxtGH/BSSeg}.
\end{abstract}



	\section{Introduction}
\begin{figure}[!t]
	\centering
	\includegraphics[scale=0.33]{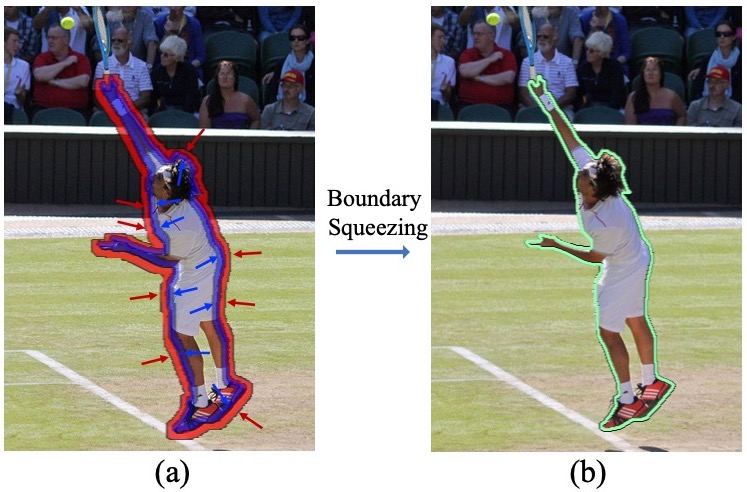}
	\caption{Illustration of the boundary squeezing process. In (a), the red arrows mean squeeze boundary features from the object's outside~(contraction process), while the blue arrows mean squeeze boundary features from inside of the object~(expansion process). These two processes complement each other and help generate accurate boundary~(green line in (b)). Best viewed on screen and zoom in.}
	\label{fig:teaser1}
	\vspace{-5mm}
\end{figure}

\begin{figure*}[!t]
	\centering
	\includegraphics[width=1.0\linewidth]{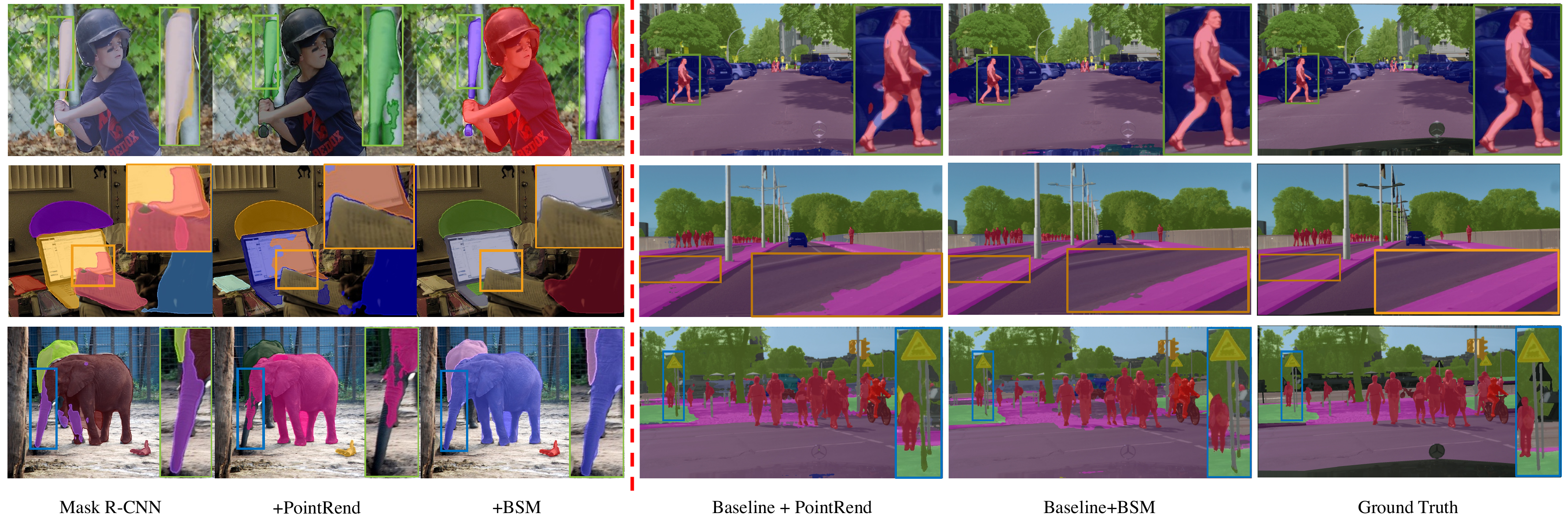}
	\caption{ Example result pairs: (a) Left: from Mask R-CNN~\cite{MaskR-CNN} vs. PointRend~\cite{PointRend} and vs. Our Boundary Squeeze~(BSM), using ResNet-50~\cite{ResNet} backbone. (b) Right: from DeeplabV3+~\cite{deeplabv3p}  PointRend~\cite{PointRend} vs. Our Boundary Squeeze. Note that our model predicts masks with more structural object boundaries where the PointRend fails to predict the finer details. }
	\label{fig:teaser_2}
\end{figure*}
The tasks of image segmentation aim to comprehensively understand the image content by densely predicting binary masks for each object (instance segmentation) or each category (semantic segmentation)~\cite{fcn,deeplabv3,MaskR-CNN}. For humans, however, image segmentation is precisely annotated in another paradigm, where the efforts are also focused on object boundaries. 
For the human, on the one hand, given the precise boundaries, one can find the masks precisely. On the other hand, given the masks, one can obtain the boundaries. For instance segmentation, the mask is represented as a binary foreground vs. background map of things where the boundaries belong to both foreground and background. In semantic segmentation, the mask is described as a binary segmentation map with a unique semantic label for the scene where the boundary lies on the intersection of different semantic maps. From this perspective, both tasks can be unified by the connection of boundaries. The finer boundaries can provide better localization performance and make the object masks more distinct and clear. The above observations motivate us in this paper to improve image segmentation by emphasizing precise localization of mask boundaries as human labeling.

Recent state-of-the-art works for both instance segmentation and semantic segmentation are developed on the fully convolution network (FCN)~\cite{fcn,PSPNet,deeplabv3p,MaskR-CNN,PointRend}. With FCN, features are densely generated on a spatial grid, where mask predictions are accomplished by classifying features in a Region of Interest~(RoI) or the whole image. These methods treat all pixels equally, and ignore the challenge of generating discriminative features for boundary pixels. Accordingly, non-uniform representations~\cite{BMaskR-CNN,PointRend} are proposed to process boundary pixels in a special way. Boundary-Preserving Mask R-CNN~\cite{BMaskR-CNN} adds a boundary branch to refine the mask, while PointRend~\cite{PointRend} adaptively selects a non-uniform set of points to refine the coarse mask during the upsampling process.

However, both works have several shortcomings. For the former, directly learning the object boundaries is hard. 
For the latter, point-based sampling and rendering along the boundary can not guarantee the structural failure case of objects. As shown in the first row in Fig.~\ref{fig:teaser_2}, the baseball bat has a similar appearance to the background wall. The rendering module in PointRend fails to classify the background and foreground things since there are no boundary bounds, and the points for rendering are sampled from coarse features without guidance.

As mentioned previously, in this work, we try to unify the two different segmentation problems via the boundary bound. Inspired by the previous traditional binary segmentation methods, such as differential edge detection and morphological processing~\cite{evans2006morphological,papari2011edge}, a new Boundary Squeeze module~(BSM) is proposed. When designing this module, our insight is squeezing the object boundary from two different directions: one from the objects' outer parts and the other from the inner parts. This strategy leads to a much finer boundary prediction. The illustration of the boundary squeezing process is shown in Fig.~\ref{fig:teaser1}. To achieve these squeezing processes, in the BSM, a flow-based warping module is proposed to let the network learn in a data-driven manner, and we term it Squeezed Feature Generator~(SFG). In BSM, there are two independent SFG supervised by two different mask labels. One label is the dilation~(or contraction) results of the original masks, while the other is the erosion~(or expansion) results of the original masks. These dual supervision signals work complementary and result in much more satisfactory boundary feature representations, contributing to accurate boundary predictions.

The Boundary Squeeze module is a general module that allows many implementations on different frameworks. It can be a plug-in module for the current state-of-the-art network in an end-to-end manner.  For example, BSM can be inserted into some Mask R-CNN-like networks'~\cite{MaskR-CNN,CascadeRCNN} mask heads in the instance segmentation task. It can also be appended at the end of modern semantic segmentation networks such as DeeplabV3+~\cite{deeplabv3p} for refining the semantic segmentation masks. These will be detailed in the experiment section. 

Our Boundary Squeeze module is evaluated on the instance and semantic segmentation tasks using COCO~\cite{COCO}, Cityscapes~\cite{Cityscapes}, LVIS~\cite{gupta2019lvis}, BDD~\cite{BDD} datasets. \textit{Qualitative}, our BSM outputs sharp and more structural boundaries and obtains significant improvement on mask quality. Compare with the previous rendering-based approach~\cite{PointRend}, our method can preserve more structural information and avoid the shortcomings of randomly selected points.  The segmentation results of our method are shown in the third and fifth columns of Fig.~\ref{fig:teaser_2}, which is much more accurate than Mask R-CNN~\cite{MaskR-CNN} and PointRend~\cite{PointRend}. \textit{Quantitative}, our method has some improvements over the PointRend with various metrics, including standard intersection-over-union based metrics for these tasks~(mask AP and mIoU).
Since these metrics are biased towards object-interior pixels and are relatively insensitive to boundary improvements~\cite{boundaryIoU}, we also compute the boundary-aware metrics~(F-score~\cite{gatedSCNN} and recent proposed boundary AP~\cite{boundaryIoU}). Our method also achieves improvement over PointRend under the same setting~\cite{detectron2} while running more efficiently. These results prove the efficiency and effectiveness of our Boundary Squeeze. To summarize, our contributions have the following aspects:
\begin{itemize}
    \item A novel and efficient module named Boundary Squeeze is designed to treat the image segmentation tasks as a boundary squeezing process. This model is differentiable and supervised via different labels obtained from the original mask annotation for free.
    \item Boundary Squeeze module can be performed as a plug-and-play module by easily being deployed into some current state-of-the-art segmentation methods, including Mask R-CNN~\cite{MaskR-CNN} and DeeplabV3+~\cite{deeplabv3p}. With our module, performances of these methods on different datasets can be significantly improved.
    \item Our approach produces finer boundary results, and better segmentation masks than some previous instance segmentation methods over different metrics on three datasets. At the same time, our model runs more efficiently. Thus, it can achieves better speed and accuracy trade-off.
    \item The proposed approach is also verified for the semantic segmentation task on Cityscapes~\cite{Cityscapes} and BDD~\cite{BDD} datasets. It also improves strong DeeplabV3+ models~\cite{deeplabv3p} by a significant margin. 
\end{itemize}

	\section{Related Work}
\noindent
\textbf{Boundary Processing:} Modeling boundary has a long history in the computer vision community. Boundary detection has been a fundamental computer vision task as it provides an essential clue for object recognition. In the era of deep learning, some CNN-based methods have significantly pushed the development of this field, such as~\cite{holistically, maninis2017convolutional, kokkinos2015pushing, he2015delving, deng2018learning, deepcontour}. On top of that, Snake~\cite{snakes} and Deep Snake~\cite{deepsnake} refine the initial object contours recursively 
through some specific operations such as circular convolution. CASENet~\cite{casenet} proposes a challenging task of category-aware boundary detection. InstanceCut~\cite{instancecut} adopts boundaries to partition semantic segmentation into instance-level segmentation. BCNet~\cite{ke2021bcnet} utilizes a bilayer convolutional network to explicitly model of occlusion relationship
via boundary information. All the above works require the model to predict the object's boundary directly. However, in some situations, such as the scene is complicated illustrated in the Fig.~\ref{fig:coco_visualize}(b) directly predicting the object's boundary is not a trivial task. 

\noindent
\textbf{Instance Segmentation:} Instance segmentation aims to detect and segment each instance~\cite{dai2016instance,hariharan2014simultaneous} in the images. The two-stage pipeline Mask R-CNN and its variants~\cite{MaskR-CNN,MSRCNN,htc,CascadeRCNN, forestrcnn, song2018cumulative} first generate object proposals using Region Proposal Network (RPN)~\cite{fasterRCNN} and then predict boxes and masks on each RoI feature. Further improvements have been made to boost its accuracy. PANet~\cite{PAN} introduces a bottom-up path to enrich the FPN features, and Mask Scoring R-CNN~\cite{MSRCNN} addresses the misalignment between the confidence score and localization accuracy of predicted masks. HTC~\cite{htc} extends the Cascade Mask R-CNN~\cite{CascadeRCNN} by augmenting a semantic segmentation branch. SCNet~\cite{scnet} balances the IoU distribution of the samples for both training and inference. Several single-stage methods~\cite{polarmask,tensormask,CondInst,blendmask,MEInst,yolact,solo,solov2} achieve significant progress and comparable results with two-stage pipelines. Meanwhile, several bottom-up approaches~\cite{neven2019instance,de2017semantic,liu2018affinity} first predict the mask of the whole image then design some methods to group the pixels belonging to the same object.

 Accurate boundary localization can explicitly contribute to the mask prediction for instance segmentation, and there are many recent work~\cite{PointRend,BMaskR-CNN,Supervised_edge_net} focus on boundary modeling. In particular, PointRend~\cite{PointRend} first predicts a low-resolution coarse mask for each instance then upsample it gradually. PointRend handles upsampling procedure as a rendering process via a shared multiple-layer network. Boundary preserving Mask R-CNN~\cite{BMaskR-CNN} proposes to predict instance-level boundaries to augment the mask head. Although those work achieves better segmentation results, the extra computation results in slow inference speed, detailed in experiment section. However, in practical usage, efficiency is also an important indicator. Therefore, when designing the model, its complexity should be taken into consideration.

\noindent
\textbf{Semantic Segmentation:} Semantic segmentation is required to assign a semantic label for each pixel. Fully convolutional networks~\cite{fcn} are the foundation of modern semantic segmentation approaches. Recent approaches try to overcome the limited receptive field of FCNs by multi-scale pooling~\cite{PSPNet,upernet}, dilated convolution~\cite{dilation,deeplabv2,deeplabv3,deeplabv3p}, non-local-like operators~\cite{nonlocal,DANet,EMANet,CCNet,dgmn,ocrnet,SGR_gcn,Li2019GlobalAT}. There are also several works on modeling boundary for semantic segmentation~\cite{boundaries_network_fields,cnn_random_wark,aaf,Task_edge_detection,gatedSCNN,decouple,he2021enhanced}. Previous works obtain better boundary localization by structure modeling, such as boundary neural fields~\cite{boundaries_network_fields}, affinity field~\cite{aaf}, random walk~\cite{cnn_random_wark}. The work~\cite{Task_edge_detection,PGN_net} adopts edge information to refine network output by predicting edge maps from intermediate CNN layers. Zhu et al.~\cite{video_propagation} uses boundary relation loss to utilize coarse predicted segmentation labels for data augmentation. DecoupleSegNet~\cite{decouple} supervises the edge and non-edge parts simultaneously. Gated-SCNN~\cite{gatedSCNN} adds a boundary stream to learn detailed low-level information by the gated convolution. Some of the methods mentioned above that use boundary information have two disadvantages: 1) directly predict the boundary may not very accurate; 2) some models need specific human design, thus, are not flexible.
	\section{Method}

\noindent
\textbf{Overview:} In this section, the motivation of our approach will be first described. Then some notations of our method will be introduced. After that, we will present the details of our Boundary Squeeze Module in which the instance segmentation setting is used for the illustration. Then the supervision signals of our model will be offered. The finally part will describe how to deploy our module on two image segmentation tasks: instance segmentation and semantic segmentation.

\begin{figure*}[!t]
	\centering
	\includegraphics[width=1.0\linewidth]{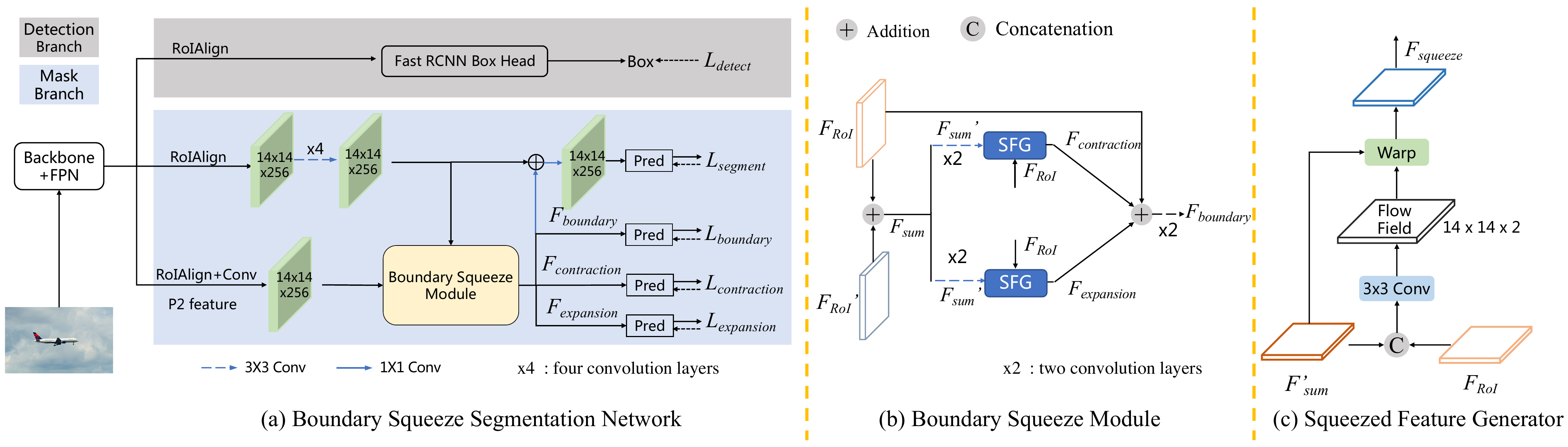}
	\caption{Network Architecture of our proposed Boundary Squeeze. (a) The entire pipeline of our Boundary Squeeze network with Mask R-CNN as the example; (b) The details of our proposed Boundary Squeeze Module. SFG: Squeeze Feature Generator; (c) The details of Squeezed Feature Generator. Best view it in color and zoom in. }
	\label{fig:network_architecture}
\end{figure*}

\subsection{Boundary Squeeze Module (BSM)}

\noindent
\textbf{Motivation:} As mentioned in the introduction section, our method is inspired by the traditional morphological processing~\cite{evans2006morphological,papari2011edge}. The boundary squeezing process is the simulation of the dilation and erosion processes via a learned neural network. Rather than the regular processing of the binary image, our method processes the image's feature map generated by the backbone network, which makes the entire process in an end-to-end manner, and such a network can be trained in a data-driven way. The second advantage of the processing feature is that this design is general and flexible so can be applied in different segmentation tasks, such as semantic segmentation and instance segmentation. 

\noindent
\textbf{Notation:} In this part, the Mask R-CNN-based instance segmentation setting will be used to illustrate some notations of our method. $F_{RoI}$ is produced using RoIAlign~\cite{MaskR-CNN} from P2-P5 FPN~\cite{fpn} features. Besides, our approach also employs the RoIAlign and P2 FPN features to produce $F_{RoI}^{'}$.
Note that the spatial size of $F_{RoI}$ is 14 $\times$ 14, while the original spatial size of $F_{RoI}^{'}$ is 28 $\times$ 28 then one 1 $\times$ 1 convolution layer is used to downsample the resolution of $F_{RoI}^{'}$ to 14 $\times$ 14. The features for the dilation~(or called \textbf{contraction}) branch and the erosion~(or called \textbf{expansion}) branch are denoted as $F_{contraction}$ and $F_{expansion}$, respectively. These two squeezed features are supervised by two different losses: $L_{contraction}$ and $L_{expansion}$. 
The squeezed feature of the boundary branch is denoted as $F_{boundary}$, and the boundary branch is supervised by $L_{boundary}$. Finally, the ground truths of the segmentation branch, boundary branch, contraction branch, and expansion branch are $G_s$, $G_b$, $G_c$, and $G_e$, separately.

\noindent
\textbf{Details of Boundary Squeeze Module:} As shown in Fig.~\ref{fig:network_architecture}(b), the BSM takes two distinct RoI features ($F_{RoI}$ and $F_{RoI}^{'}$) as inputs and outputs three different features, including $F_{boundary}$, $F_{contraction}$, and $F_{expansion}$. 
First, $F_{RoI}$ and $F_{RoI}^{'}$ are added together to get $F_{sum}$. The choice of $F_{RoI}^{'}$ follows the design of the previous works~\cite{deeplabv3p,PointRend,blendmask},  by introducing low-level and high-frequency details into the $F_{RoI}$, which benefits the boundary squeezing process. Then $F_{sum}$ is fed into two parallel branches: contraction branch and expansion branch. In both branches, two 3 $\times$ 3 convolution layers are first used to generate its unique feature $F_{sum}^{'}$. After that, a Squeezed Feature Generator~(SFG) is utilized to produce the squeezed feature~($F_{contraction}$ or $F_{expansion}$). $F_{contraction}$ and $F_{expansion}$ are used to squeeze boundary features from the outside and inside of an object.
SFG will be detailed in the following parts. Finally, both squeezed features and $F_{RoI}$ are summed together in a residual manner~\cite{ResNet} as the final squeezed boundary feature $F_{boundary}$, which is formulated as:
\begin{equation}
	F_{boundary} = F_{contraction} + F_{expansion} + F_{RoI}.
	\label{equ:fc_fe_fr}
\end{equation}

\noindent
\textbf{Squeezed Feature Generator:} We propose a flow-based approach to generate the squeezed features. The inputs of SFG are $F_{sum}^{'}$ and $F_{RoI}$, and the output is one squeezed feature~(\ie~$F_{contraction}$ or $F_{expansion}$). Specifically, 
following~\cite{FlowNet}, the squeezed feature generator first concatenates $F_{sum}^{'}$ and $F_{RoI}$ then adopts one $3 \times 3$ convolution layer to generate the flow field $\delta \in \RR^{14 \times 14 \times 2}$ for each RoI. 
With $\delta$, features of each position $\pp_{l}$ on the standard spatial grid $F_{sum}^{'}$ are mapped to a new position $\hat{\pp}$ via $\pp_{l}+\delta_{l}(\pp_{l})$. The differentiable bilinear sampling mechanism is used to achieve this process. This sampling mechanism, proposed in the spatial transformer networks~\cite{STN,DFF}, linearly interpolates the values of the four nearest neighbor pixels of $\pp_{l}$ from $F_{sum}^{'}$. This process can be formulated as: 

\begin{equation}
F_{squeeze}(p_{x}) = \sum_{\pp \in \mathcal{N}(\pp_{l})} w_\pp F_{sum}^{'}(\pp),
\label{equ:warpping}
\end{equation}
where $w_\pp$ calculated from flow map $\delta$, represents bilinear kernel weights on the warped spatial grid. $\mathcal{N}$ represents the number of involved neighboring pixels. The whole pipeline of SFG is illustrated in Fig.~\ref{fig:network_architecture}(c).

Essentially, the warping process is the deformation of the entire RoI feature. Accordingly, one can also choose other methods to achieve the same purpose, like deformable convolution~\cite{dcn}. However, though deformable convolution results in a similar performance, it needs more inference time, relevant results can be found in the experiment section. 

\begin{figure}[!t]
	\centering
	\includegraphics[scale=0.93]{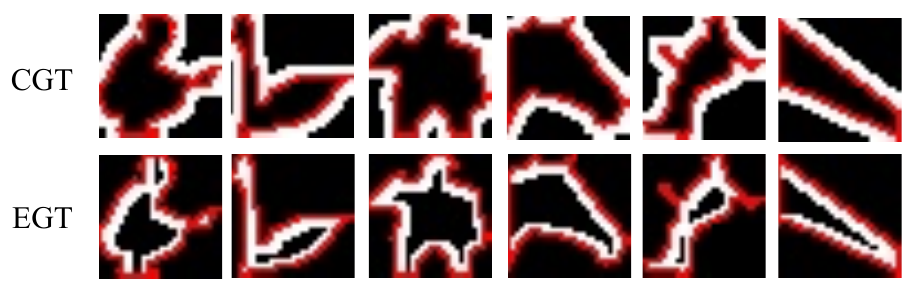}
	\caption{ Visualization of Contraction branch Ground Truth~(CGT) and Expansion branch Ground Truth~(EGT). White pixels are positive while black pixels are negative. Boundary pixels are labeled as red to distinguish positive pixels and negative pixels.}
	\label{fig:ce_ground_truth}
\end{figure}

\noindent
\textbf{Supervision Signals:} The different supervision signals control the warping operation's direction for the contraction branch and expansion branch. Given a binary mask annotation $G_s$, our method uses Equ.~(\ref{equ:contraction_expansion_gt}) to generate $G_c$ and $G_e$ where $H_{dilation}$ is the dilation operation in morphological processing that can extend the $G_s$'s boundary into the background, and $H_{erosion}$ is the erosion operation used to shrink $G_s$ from its original border. $K$ is the kernel size of $H_{dilation}$ and $H_{erosion}$:
\begin{equation}
\left\{
             \begin{array}{lr}
             G_{c} = H_{dilation}(G_s, K) - G_{s} &  \\
             G_{e} = G_{s} - H_{erosion}(G_s, K).&  
             \end{array}
\right.
\label{equ:contraction_expansion_gt}
\end{equation}
The visualization of some $G_c$ and $G_e$ can be found in Fig.~\ref{fig:ce_ground_truth}. 
Following~\cite{BMaskR-CNN}, the Laplacian operator is used to generate $G_b$. Finally, the supervision signal of the segmentation branch is the original $G_s$.

\begin{figure}[!t]
	\centering
	\includegraphics[scale=0.61]{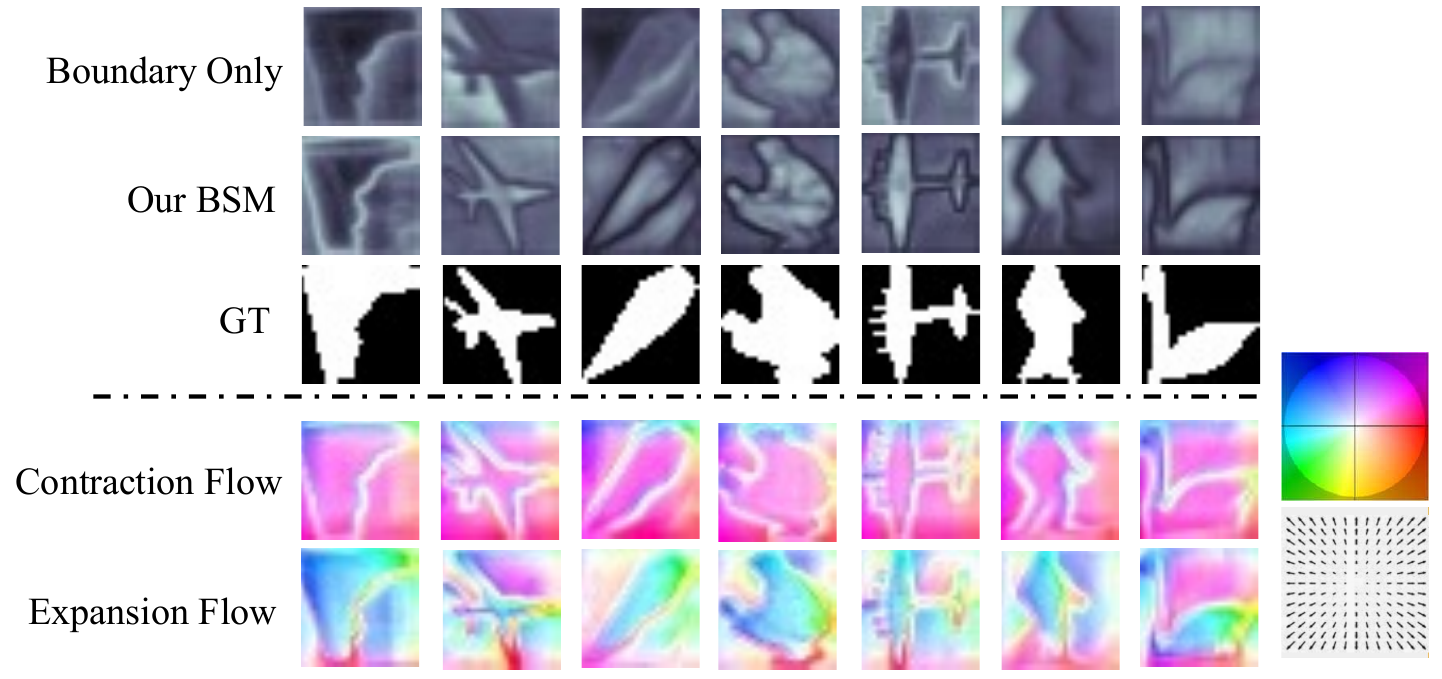}
	\caption{ Visual Interpretation of Our proposed BSM. (a) The top sub-figure shows the comparison results on instance feature representation of boundary. (Boundary Only supervision and our BSM results). GT means the Ground Truth mask. Our BSM outputs thinner and  more precise boundaries, and it also contains more structural information. 
	(b) The bottom sub-figure visualizes the two types of learned flow fields. Best view it on the screen and zoom in.}
	\label{fig:visual_interpretation}
\end{figure}

\noindent
\textbf{Visual Interpretation:} To better understand our boundary squeezing process, we exhibit two types of visual interpretations in Fig.~\ref{fig:visual_interpretation}. The top part of Fig.~\ref{fig:visual_interpretation} is the instance-aware feature visualization, and the bottom part is the visualization of two flow fields learned from two independent branches.
As shown at the top of Fig.~\ref{fig:visual_interpretation}, compared with directly predict boundary, like~\cite{boundaries_network_fields} does, our BSM produces a clearer and thinner boundary feature representation $F_{boundary}$. Note that the Principal Component Analysis~(PCA)~\cite{wold1987principal} is utilized to reduce the number of $F_{boundary}$'s dimensions into three for visualization. This comparison result proves our motivation: squeezed boundary feature results in a more discriminative and structural-preserving representation.
The bottom parts of Fig.~\ref{fig:visual_interpretation} show two different flow fields. For the contraction branch, the direction of learned flow points is to the inner parts of instance objects, while the direction of learned flow mainly expands to the outside of instance objects in the expansion branch. Both flows squeeze the object boundary in a complementary way and benefit each other, proven in the experiment section.

\subsection{Boundary Squeeze on the Image Segmentation Tasks}

This section will describe how to deploy our proposed BSM on two image segmentation tasks. 

\noindent
\textbf{Instance Segmentation:} For instance segmentation task, the Mask R-CNN~\cite{MaskR-CNN} is adopted as the baseline method. As shown in Fig.~\ref{fig:network_architecture}(a), the proposed BSM is inserted after the four consecutive 3 $\times$ 3 convolution layers of Mask R-CNN head. And the final prediction head~\footnote{One deconvolution layer with 2 $\times$ 2 kernel size, one ReLU layer, and one 1 $\times$ 1 convolution layer.}~(denote as Pred in Fig.~\ref{fig:network_architecture}(a)) for every branch is the same with Mask R-CNN. The kernel size of dilation and erosion operations is set to 5, and the related experiments can be found in the experiment section. The total loss $L_{instance}$ is the combination of multi-tasks learning, formulated as Equ.~(\ref{equ:d_sum_s}):
\begin{equation}
	L_{instance}=L_{detection}+L_{mask},
	\label{equ:d_sum_s}
\end{equation}
  where $L_{detection}$ is the loss of detection branch and $L_{mask}$ can be expressed by Equ.~(\ref{equ:sum_mask}):
 \begin{equation}
 \begin{split}
 	L_{mask} = L_{segmentation} + L_{boundary} \\+ L_{contraction} + L_{expansion},
 	\label{equ:sum_mask}
 \end{split}
 \end{equation}
   where $L_{segmentation}$ is Binary Cross-Entropy loss~(BCE), $L_{boundary}$ is the combination of BCE loss and Dice loss~\cite{DiceLoss}, both $L_{contraction}$ and $L_{expansion}$ are Dice loss. During inference, the output of the segmentation branch is served as the model's output.


\noindent
\textbf{Semantic Segmentation:} For the semantic segmentation task, the DeeplabV3+~\cite{deeplabv3p} is chosen as our baseline. In particular, the proposed BSM is appended at the end of the DeeplabV3+ head~(after the output of Atrous Spatial Pyramid Pooling, ASPP). Our BSM treats the entire feature map as one RoI in instance segmentation task. The low-level feature is the same as the DeeplabV3+ design. For supervision signal generation, since most semantic segmentation datasets~\cite{Cityscapes,ADE20K,BDD} have no separate background class, the binary setting for each class is adopted. Specially, each semantic class is treated as one binary mask, which means each class in the semantic segmentation label map is one RoI in the instance segmentation setting to simulate the expansion and contraction processes. Then our BSM can fully utilize the same setting as instance segmentation. The kernel size of dilation and erosion operations is set to 15. Entire training process is supervised by both original Cross-Entropy loss and three extra losses (boundary loss and two independent Dice losses on two squeezing branches), which are the same as the instance segmentation task. 
	\section{Experiment}
\subsection{Instance Segmentation}

\noindent
\textbf{Overview:}
We first conduct experiments on the instance segmentation task. Extensive ablation experiments are performed on the challenging COCO dataset~\cite{COCO}. Besides, experimental results on the Cityscapes~\cite{Cityscapes} and LVIS~\cite{gupta2019lvis} are also presented when comparing our model with other approaches.

\begin{table*}[!t]
	\begin{minipage}[!t]{\linewidth}
	    \centering
		\begin{minipage}{.48\linewidth}
		\subfloat[Effect of each branch in BSM]{
		\resizebox{0.96\textwidth}{!}{%
		\centering
		\begin{tabular}{c c c | l l l}
		\hline
			Boundary & Contraction & Expansion & AP & AP$_{50}$ & AP$_{75}$ \\  
		\hline
			- & - & - & 35.2 & 56.2 & 37.6 \\ 
			\checkmark & - & - & 36.0 & 56.4 & 38.7 \\ 
			-& \checkmark & - & 35.8 & 56.6 & 38.4 \\
			-& - & \checkmark & 36.1 & 56.8 & 38.8 \\
			-& \checkmark & \checkmark & 36.5 & 56.8 & 39.2 \\
			\checkmark &\checkmark &\checkmark & \textbf{36.9} & \textbf{57.2} & \textbf{39.9} \\
			\hline
		\end{tabular}
				}}
		\end{minipage}
	    \begin{minipage}{.46\linewidth}
		\subfloat[Contraction and expansion branches' loss function]{
		\resizebox{0.96\textwidth}{!}{%
		\centering
		\begin{tabular}{c c c |l l l}
		\hline
			BCE & Weighted BCE & Dice & AP & AP$_{50}$ & AP$_{75}$ \\  
		\hline
			\checkmark & - & - & 36.2 & 56.5 & 38.9 \\ 
			- & \checkmark & - & 36.6 & 57.1 & 39.6  \\
			- & - & \checkmark & \textbf{36.9} & \textbf{57.2} & \textbf{39.9}  \\
			\checkmark & - & \checkmark & 36.6 & \textbf{57.2}  & 39.6\\
			- &\checkmark &\checkmark & 36.5 & 56.8 & 39.4\\
			\hline
		\end{tabular}	
				}}
		\end{minipage}
		\begin{minipage}{.41\linewidth}
		\subfloat[Effectiveness of adding low-level RoI features]{
			\footnotesize
			\resizebox{0.86\textwidth}{!}{%
			\begin{tabular}{c|l l l}
			\hline
				P2 RoI features & AP & AP$_{50}$ & AP$_{75}$  \\
				\hline
				w/o &  36.5 & 56.8 & 39.3\\
				\hline 
				w & \textbf{36.9} & \textbf{57.2} & \textbf{39.9} \\
				\hline
		\end{tabular}}}
		\end{minipage}
		\begin{minipage}{.52\linewidth}
		\subfloat[Ablation of feature information fusion]
		{
			\footnotesize
			\resizebox{0.92\textwidth}{!}{%
			\begin{tabular}{c c|l l l}
			\hline
				Add Mask feature & Boudary2Mask conv & AP & AP$_{50}$ & AP$_{75}$  \\
				\hline
				- & - & 36.0 & 56.7 & 38.6 \\
				\checkmark & - & 36.6 & 57.0 & 39.0 \\
				- & \checkmark & 36.4 & 56.5 & 39.5 \\
				\checkmark & \checkmark & \textbf{36.9} & \textbf{57.2} & \textbf{39.9} \\
				\hline
		\end{tabular}}}
		\end{minipage}
		\begin{minipage}{.48\linewidth}
		\subfloat[Comparison of different feature deformation methods]
		{
			\footnotesize
			\resizebox{0.98\textwidth}{!}{%
			\begin{tabular}{c c c c | c}
			\hline
				Method & AP & AP$_{50}$ & AP$_{75}$ & Inference Time  \\
				\hline
				w/o fusion & 36.0 & 56.5 & 37.7 & 52ms \\
				w DGMN~\cite{dgmn} & 36.6 & 57.0 & 39.2 & 60ms \\
				w DCNV2~\cite{dcnv2} & 36.8 & 57.2 & 39.4 & 62ms \\
				w ours & 36.9 & 57.2 & 39.9 & 54ms \\
				\hline
		\end{tabular}}}
		\end{minipage}
		\hspace{2mm}
		\begin{minipage}{.46\linewidth}
		\subfloat[Results of using different backbone network]
		{
			\footnotesize
			\resizebox{0.96\textwidth}{!}{%
			\begin{tabular}{c c|l l l}
			\hline
				Method & Backbone & AP & AP$_{50}$ & AP$_{75}$  \\
				\hline
				Baseline & R-101-FPN & 37.1 & 58.4 & 39.9 \\
				+Ours & R-101-FPN & 38.6 & 59.3 & 41.7 \\
				Baseline & X-101-FPN & 39.0 & 61.3 & 41.8 \\
				+Ours & X-101-FPN & 40.6 & 62.1 & 43.8 \\
				\hline
		\end{tabular}}}
		\end{minipage}
	
	\end{minipage}
	\caption{ \textbf{Ablation studies.} We first verify the effectiveness of each branch of our model in (a); then results of some experiments with different loss functions for contraction and expansion branches are provided in (b); after that, (c) explores the influence of low-level RoI features; and the impact of information fusion of different branches is investigated in (d); (e) offers the results using different feature deformation methods; (f) is used to show the generality of our method where R means ResNet and X means ResNeXt~\cite{xie2017aggregated}. All models are trained on COCO \textit{train2017} and tested on COCO \textit{val2017}. Inference times are tested using one V100 GPU with single scale testing.
	}\label{tab:ablations}
\end{table*}

\noindent 
\textbf{Dataset:}
\label{sec:instance_segmentation_datasets_and_metrics}
\textbf{(a) COCO} contains 118k images for training, 5k images for validation, and 20k images for testing. Following the common practice~\cite{MaskR-CNN, BMaskR-CNN, focalloss}, our models are trained using the training set~(\textit{train2017}) and tested results on the validation set~(\textit{val2017}) for ablation studies. Results on the test set~(\textit{test-dev2017}) are also reported for comparing with other instance segmentation methods. \textbf{(b) Cityscapes} has 2975 images, 500 images, and 1525 images for training, validation, and test, respectively. For instance segmentation task, Cityscapes has 8 instance categories. \textbf{(c) LVIS} is a recently proposed dataset with the same images as COCO but has more than 1K categories. The annotations' quality of Cityscapes and LVIS are much higher than COCO, especially on the boundary.

\noindent \textbf{Metrics:} The standard mask AP is used as the primary metric. To better evaluate the boundary quality of the instance segmentation results, the newly proposed boundary AP~\cite{boundaryIoU} is reported in some experiments.

\noindent
\textbf{Implementation details:}
The PyTorch library~\cite{pytorch} and Detectron2~\cite{detectron2} codebase are employed to implement all the models. Our model uses the ResNet-50 pre-trained on ImageNet~\cite{imagenet} with FPN~\cite{fpn} as the backbone network unless otherwise stated. All models are trained using 8 GPUs with stochastic gradient descent~(SGD). Momentum and weight decay are set as 0.9 and 0.0001, respectively. For the COCO dataset, the training images are resized to a shorter side from 640 to 800 with a step of 32 pixels and their longer side less or equal to 1333. At inference time, images are resized to the short side of 800 pixels. In the ablation studies, all models are trained for 90k iterations. The learning rate is initialized to 0.02 and reduced by a factor of 10 at iteration 60k and 80k. A mini-batch contains 16 images~(2 images per GPU). 

For the Cityscapes dataset, 8 images~(1 image per GPU) are used for one mini-batch. All models are trained on the training set for 24k iterations with the learning rate initialized to 0.01, then reduced to 0.001 at the iteration 18k. All models are tested on the validation set. At training time, images are resized randomly to a shorter edge from 800 to 1024 pixels with a step of 32 pixels, and their longer edge is less or equal to 2048 pixels. At inference time, images are resized to the shorted edge of 1024 pixels. 

For the LVIS dataset, the LVIS$^*$\scriptsize{v0.5} \normalsize version is adopted. Following~\cite{PointRend}, the LVIS$^*$\scriptsize{v0.5} \normalsize dataset is constructed by keeping only the 80 COCO categories from LVIS\scriptsize{v0.5}. \normalsize All models are trained on COCO $train2017$ set with standard 1$\times$ schedule~\cite{detectron2} and tested on the validation set of LVIS$^*$\scriptsize{v0.5} \normalsize where images are resized to the short edge of 800 pixels.
\subsubsection{Ablation studies}
\ 

\noindent \textbf{Each branch in BSM:} In Tab.~\ref{tab:ablations}(a), the effectiveness of each branch of BSM is first verified, and Mask R-CNN~\cite{MaskR-CNN} is used as the baseline. After adding boundary branch, contraction branch, and expansion branch separately, AP is improved by 0.8, 0.6, and 0.9, respectively. These results prove the effectiveness of each branch. Adding both contraction and expansion branches leads to 1.3 AP gain, which means these two branches can complement each other. Adding all the three branches results in another 0.4 AP gain, which indicates the effectiveness of using boundary supervision explicitly.

\noindent \textbf{Loss function design of contraction and expansion branch:} Then the performance of the choices with loss function on contraction and expansion branches are demonstrated in Tab.~\ref{tab:ablations}(b). Using Binary Cross-Entropy~(BCE), weighted Binary Cross-Entropy~(weighted BCE), Dice loss~\cite{DiceLoss}, severally, the final AP is 36.2, 36.6, 36.9, respectively. The latter two loss functions achieve better performance since they can better handle the class imbalance problems in the contraction and expansion branches. After combining Dice loss with BCE or weighted BCE, there are slight performance drops, partially because the Cross-Entropy loss focuses more on pixel-level difference~\cite{deng2018learning}, while contraction and expansion branches need to concentrate more on the similarity of two sets of image pixels. Therefore, the Dice loss is adopted as the loss function for contraction and expansion branches.

\noindent \textbf{Low-level RoI features:} The importance of adding low-level RoI features $F_{RoI}^{'}$ is investigated in the ablation study of Tab.~\ref{tab:ablations}(c). Low-level features can provide detailed positional information to improve the segmentation quality proven in previous work~\cite{deeplabv3p}. As shown in Tab.~\ref{tab:ablations}(c), using low-level RoI features in our approach can contribute to 0.4 AP gain.

\noindent \textbf{Information fusion among different features:} The effectiveness of information fusion among different branches' features is further explored. Tab.~\ref{tab:ablations}(d) offers some results of this aspect: adding mask features to the boundary branch in a residual connection manner results in 0.6 AP gain, which indicates this process can produce better feature representation for boundary prediction. Since there is some information gap between the final boundary feature and mask feature, one 1 $\times$ 1 convolution layer~(the vertical solid blue line in Fig.~\ref{fig:network_architecture}(a)) is added to our model when merging boundary feature to mask feature, and this process leads to 0.4 AP gain. After using the above two operations, our method gains 0.9 AP.

\noindent \textbf{Different feature deformation methods:} In our implementation, the feature warping strategy is adopted to generate the contraction and expansion features. In the experiments of Table.~\ref{tab:ablations}(e), some other similar methods are employed to produce these features. Our baseline model's AP is 36.0, and it does not use any feature deformation method. After using DGMN~\cite{dgmn}, there is 0.6 AP gain, but with an extra 8ms inference time. DCNV2~\cite{dcnv2} and our feature warping method achieve similar performance~(36.8 and 36.9), but DCNV2 uses more inference time. Thus, the flow-based warping is chosen in our implementation.

\noindent \textbf{Different backbone networks:} In addition to ResNet50, Some experiments on other backbone networks, including ResNet101 and ResNeXt101~\cite{xie2017aggregated}, are carried out. Results are shown in Table.~\ref{tab:ablations}(f), with the ResNet101 backbone, our method results in a gain over the baseline of 1.5 AP. The 1.6 AP gain is found of our method on the ResNeXt101 backbone. These decent results prove the generality of our approach.

\begin{table}[!t]\setlength{\tabcolsep}{9pt}
	\centering
	\begin{threeparttable}
		\scalebox{0.80}{
		\begin{tabular}{c c c c}
					\hline
					 Method  & Inference Time  & AP\scriptsize{$^{\text{mask}}$} & AP\scriptsize{$^{\text{boundary}}$} \\  
					\hline
					 Mask R-CNN~\cite{MaskR-CNN} & \textbf{48ms} & 35.2 & 21.3 \\
					 PointRend~\cite{PointRend} & 73ms & 36.2 & 23.3 \\
				     BMask R-CNN~\cite{BMaskR-CNN} & 58ms & 36.6 & 23.4\\
					 B2Inst-BlendMask~\cite{kim2021devil}  & 103ms & 36.7 & -  \\
					 BSM & 54ms & \bf{36.9} & \bf{23.6} \\
					\hline
				\end{tabular}}
		\caption{Comparison results with baseline and related methods on COCO \textit{val2017}. All the models are trained and tested using the same setting for fair comparison.}
		\label{tab:comparison_results_with_other_boundary_method}
	\end{threeparttable}
\end{table}

\begin{table*}[!h]\setlength{\tabcolsep}{7pt}
	\begin{center}
	\scalebox{1.0}{
	\begin{tabular}{l|c|c|c|c|cc|ccc}
		\hline
		Method & Backbone & Aug. & Sched. & AP & AP$_{50}$ & AP$_{75}$ & AP$_{S}$ & AP$_{M}$ & AP$_{L}$ \\
		\hline
		Mask R-CNN$^*$ ~\cite{MaskR-CNN} & R-50-FPN & & $1\times$ & 35.1 & 56.4 & 37.5 & 18.9 & 37.0 & 46.1 \\
		CondInst ~\cite{CondInst} & R-50-FPN & & $1\times$ & 35.4 & 56.4 & 37.6 & 18.4 & 37.9 & 46.9 \\
		\textbf{BSM} & R-50-FPN & & $1\times$ & \textbf{36.5} & \textbf{57.0} & \textbf{39.3} & \textbf{19.4} & \textbf{38.6} & \textbf{47.8} \\
		\hline
		TensorMask~\cite{tensormask} & R-50-FPN & \checkmark & $6\times$ & 35.4 & 57.2 & 37.3 & 16.3 & 36.8 & 49.3 \\
		BMask R-CNN$^*$~\cite{BMaskR-CNN} & R-50-FPN & \checkmark & $1\times$ & 36.7 & 57.0 & 39.4 & 17.8 & 39.1 & 49.6 \\
		CondInst~\cite{CondInst} & R-50-FPN & \checkmark & $1\times$ & 35.9 & 56.9 & 38.3 & 19.1 & 38.6 & 46.8 \\
		Mask R-CNN$^*$~\cite{MaskR-CNN}& R-50-FPN & \checkmark & $3\times$ & 37.3 & 58.9 & 40.1 & 21.1 & 39.3 & 47.8 \\
		\textbf{BSM} & R-50-FPN & \checkmark & $1\times$ & 37.0 & 57.6 & 40.0 & 20.5 & 39.2 & 47.9 \\
		\textbf{BSM} & R-50-FPN & \checkmark & $3\times$ & \textbf{38.9} & \textbf{60.0} & \textbf{42.1} & \textbf{22.0}& \textbf{41.1} & \textbf{49.8} \\
		\hline
		Mask R-CNN~\cite{MaskR-CNN} & R-101-FPN & \checkmark & $3\times$ & 38.8 & 60.9 & 41.9 & 21.8 & 41.4 & 50.5 \\
		MS R-CNN~\cite{MSRCNN} & R-101-FPN & & 18e & 38.3 & 58.8 & 41.5 & 17.8 & 40.4 & 54.4\\
		BMask R-CNN$^*$~\cite{BMaskR-CNN} & R-101-FPN & \checkmark & $1\times$ & 38.0 & 59.6 & 40.9 & 18.1 & 40.2 & \bf{54.8} \\
		YOLACT-700~\cite{yolact} & R-101-FPN & \checkmark & $4.5\times$ & 31.2 & 50.6 & 32.8 & 12.1 & 33.3 & 47.1 \\
		SipMask~\cite{sipmask} & R-101-FPN & \checkmark & 6$\times$ & 38.1 & 60.2 & 40.8 & 17.8 & 40.8 & 54.3 \\
		BlendMask~\cite{blendmask} & R101-FPN & \checkmark & 3$\times$ & 38.4 & 60.7 & 41.3 & 18.2 & 41.5 & 53.3 \\
		CondInst~\cite{CondInst} & R-101-FPN & \checkmark & 3$\times$ & 39.1 & 60.9 & 42.0 & 21.5 & 41.7 & 50.9 \\
		SOLOv2~\cite{solov2} & R-101-FPN & \checkmark & 3$\times$ & 39.7 & 61.8 & 43.1 & 21.0 & 42.2 & 53.5 \\
		BCNet~\cite{ke2021bcnet} & R-101-FPN & \checkmark & $3\times$ & 39.8 & 61.5 & 43.1 & 22.7 & 42.4 & 51.1 \\
		\textbf{BSM} & R-101-FPN & \checkmark & $1\times$ & 38.6 & 59.6 & 41.8 & 21.3 & 41.2 & 49.8 \\
		\textbf{BSM} & R-101-FPN & \checkmark & $3\times$ & \textbf{40.4} & \textbf{61.9} & \textbf{43.8} & \textbf{22.8} & \textbf{43.0} & 52.4 \\
		\textbf{BSM} & X-101-FPN & \checkmark & $3\times$ & \textbf{41.6} & \textbf{63.7} & \textbf{45.1} & \textbf{24.4} & \textbf{44.4} & 53.4 \\
		\hline
		Cascade Mask R-CNN$^*$ & R50-FPN & $\checkmark$ & 1$\times$ & 36.4 & 56.9 & 39.2 & 17.5 & 38.7 & 52.5 \\
		Cascade BMask R-CNN$^*$ & R-50-FPN & $\checkmark$ & 1$\times$ & 37.5 & 57.3 & 40.7 & 17.5 & 39.8 & 54.1 \\
		Cascade BMask R-CNN$^*$ & R-101-FPN & $\checkmark$ & 3$\times$ & 40.5 & 61.5 & 44.0 & 20.4 & 40.9 & \textbf{56.8} \\
		\textbf{Cascade BSM} & R-50-FPN & $\checkmark$ & 1$\times$ & 37.9 & 58.0 & 41.2 & 20.7 & 39.7 & 51.5\\
		\textbf{Cascade BSM} & R-101-FPN & $\checkmark$ & 3$\times$ & \textbf{41.0} & \textbf{62.0} & \textbf{44.6} & \textbf{23.3} & \textbf{41.1} & 54.0 \\
		\hline
	\end{tabular}}
	\end{center}
		\caption{ Comparisons with state-of-the-art methods on COCO \textit{test-dev2017}. All models are trained on COCO \textit{train2017}. Aug. means using multi-scale training data augmentation. Sched. means learning rate schedule during training. 1$\times$ is $90K$ iterations, 2$\times$ is $180K$ iterations, 3$\times$ is $270K$ iterations and so on, 18e means 18 epochs. R means ResNet and X means ResNeXt~\cite{xie2017aggregated}. $^*$ means our implementation.}
	\label{table:comparisons_state_of_the_art_methods}
	\vspace{2mm}
\end{table*}

\begin{figure}[!t]
	\centering
	\includegraphics[scale=0.7]{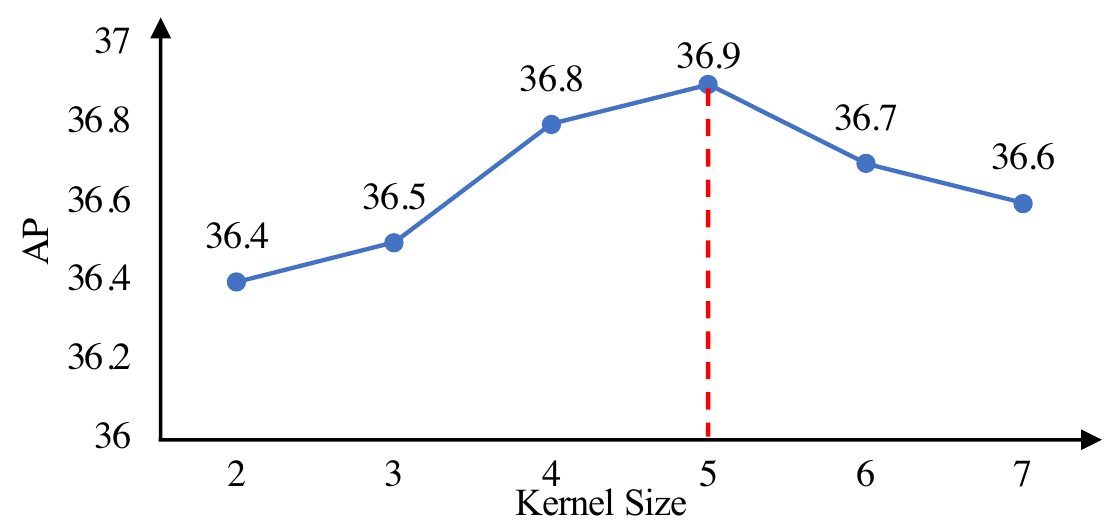}
	\caption{ Influence of kernel size. Our model achieves the best result when the kernel size equals 5.}
	\label{fig:influence_of_kernel_size_instance}
\end{figure}

\noindent \textbf{Efficiency and Effectiveness: } Both accuracy and efficiency are essential factors to consider in practical use. Therefore, the inference time, mask AP~(AP$^{\text{mask}}$), and recently proposed boundary AP~(AP$^{\text{boundary}}$) ~\cite{boundaryIoU} of our method are further compared with baseline Mask R-CNN and other related works including, PointRend~\cite{PointRend}, BMask R-CNN~\cite{BMaskR-CNN}, and B2Inst-BlendMask~\cite{kim2021devil}. Compared with AP$^{\text{mask}}$, AP$^{\text{boundary}}$ can better evaluate the accuracy of the boundary. All models employ ResNet50 as backbone and use the same inference setting except B2Inst-BlendMask$\footnote{Since the authors of B2Inst do not release their codes and models, all results of B2Inst-BlendMask come from the B2Inst paper.}$. Results are shown in Table.~\ref{tab:comparison_results_with_other_boundary_method}. Our method has the fastest inference speed among related methods and outperforms other works on both AP$^{\text{mask}}$ and AP$^{\text{boundary}}$.

The ground truths of contraction and expansion branches are generated based on the original mask ground truths, which have spatial scales of 28 $\times$ 28 in our method. The dilation and erosion operations in the classic morphological processing~\cite{evans2006morphological,papari2011edge} are used to generate them. The kernel size of these operations controls the number of positive pixels in these ground truths. As shown in Fig.~\ref{fig:influence_of_kernel_size_instance}, when the kernel size is set to 5, our model achieves the best mask AP. So the kernel size is set to 5 in other instance segmentation experiments. Theoretically, with the decrease of kernel size, both ground truths of contraction and expansion branches will be more similar to the boundary ground truth. Oppositely, with the increase of kernel size, these ground truths will be more analogous to the original mask ground truth. Both cases cannot be able to utilize the original supervision information fully.

\begin{figure*}[!t]
	\centering
	\includegraphics[scale=0.31]{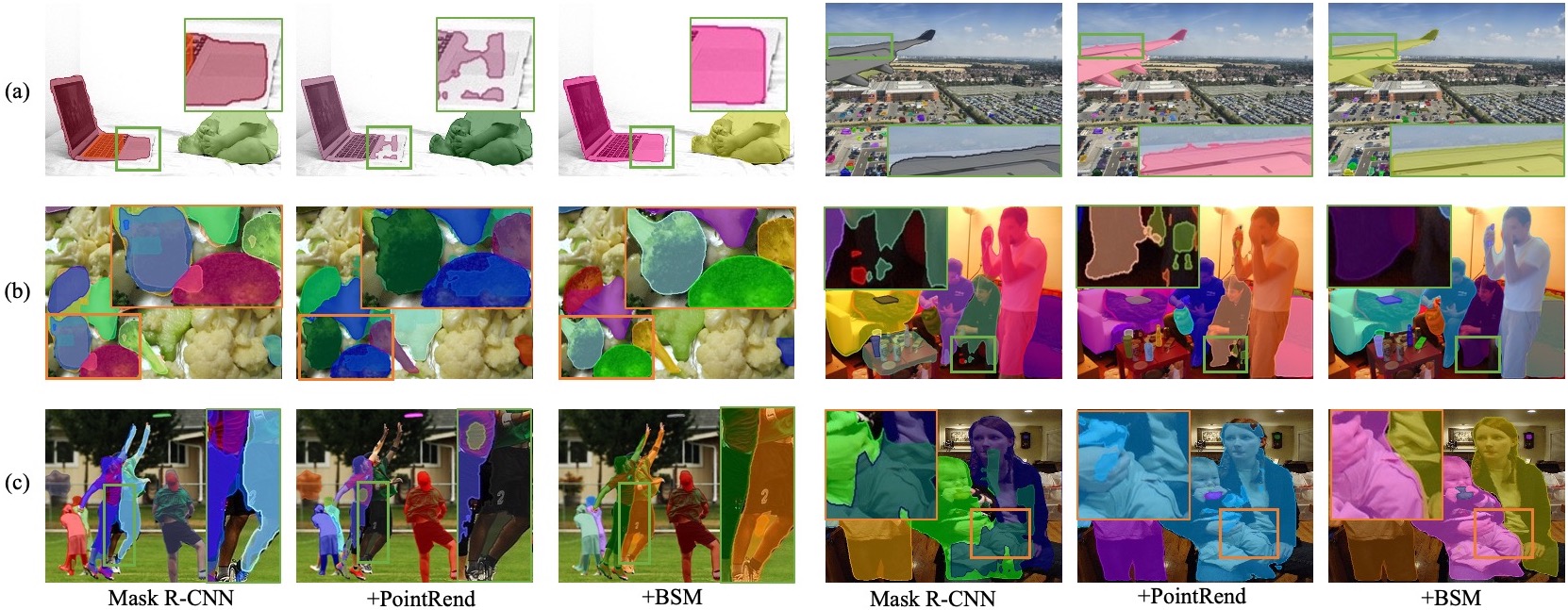}
	\caption{ Visualization and comparison results of Mask R-CNN~\cite{MaskR-CNN}, PointRend~\cite{PointRend} and our BSM on the COCO $\textit{val2017}$ using ResNet-50~\cite{ResNet} with FPN~\cite{fpn} as backbone. Compared with the other two methods, BSM can produce better segmentation results. Best viewed on screen and zoom in.}
	\label{fig:coco_visualize}
\end{figure*} 

\begin{table}[!h]\setlength{\tabcolsep}{7pt}
	\centering
	\begin{threeparttable}
		\scalebox{0.85}{
		\begin{tabular}{c| c| c c|c c}
					\hline
					 Method  & Backbone  & AP & AP$_{50}$& AP$^*$ & AP$^*_{50}$ \\  
					\hline
					 Mask R-CNN~\cite{MaskR-CNN} & R-50-FPN & 33.0 & 50.0 & 37.4 & 58.9 \\
				     PointRend~\cite{PointRend} & R-50-FPN & 35.1 & 62.6 &39.3 & 59.7 \\
					 BMask R-CNN~\cite{BMaskR-CNN} & R-50-FPN & 35.7 &61.9 &39.2 & 59.5  \\
					 BSM & R-50-FPN & \bf{36.1} & \bf{63.0}& \bf{39.7} &\bf{60.2} \\
					\hline
					 Mask R-CNN~\cite{MaskR-CNN} & R-101-FPN & 33.6 & 61.8& 39.6 &61.2 \\
				     PointRend~\cite{PointRend} & R-101-FPN & 35.8 & 63.5 &41.5 & 62.1\\
					 BMask R-CNN~\cite{BMaskR-CNN} & R-101-FPN & 36.3 & 62.7 & 41.2 & 61.9 \\
					 BSM & R-101-FPN & \bf{36.9} & \bf{63.5} & \bf{41.9} &\bf{62.4} \\
					\hline
				\end{tabular}}
		\caption{ Experiments results of BSM and another three methods on Cityscapes and LVIS datasets. AP and AP$_{50}$ are reported on the validation set of Cityscapes, while AP$^*$ and AP$^*_{50}$ are reported on the validation set of LVIS$^*$\scriptsize{v0.5}.}
		\label{tab:comparison_results_with_other_boundary_method_cityscape_lvis}
	\end{threeparttable}
\vspace{-5mm}
\end{table}
\subsubsection{Main Results} 
\

\noindent 
\textbf{Comparison with state-of-the-art methods on COCO:} In Table.~\ref{table:comparisons_state_of_the_art_methods}, BSM is compared to the state-of-the-art instance segmentation methods on COCO $\textit{test-dev2017}$. BSM employs 3$\times$ learning rate schedule with ResNet101 as the backbone network achieves 40.4 AP, which is much better than other state-of-the-art methods. When using cascade architecture, BSM also conducts state-of-the-art results.

\noindent \textbf{Visualization Results:} Two types of visualization results are provided in this part. 

First, Fig.~\ref{fig:coco_visualize} provides some visual comparison results between our model and the other two methods~(Mask R-CNN~\cite{MaskR-CNN} and PointRend~\cite{PointRend}) on the COCO dataset. Our model can generate better segmentation results than both approaches. Note that PointRend also model some points of the object boundary, it focuses on refining the segmentation results around boundary pixels. BSM can segment instances well in three challenging situations: (a) the appearance of instances is very similar to its surrounding background, (b) complicated scenes, (c) instances overlapping. We argue our model can produce better segmentation results than the other two methods because our model can generate accurate boundaries. And these precise boundaries can help locate instances and distinguish different instances well.

\begin{figure*}[t!]
	\centering
	\includegraphics[scale=0.73]{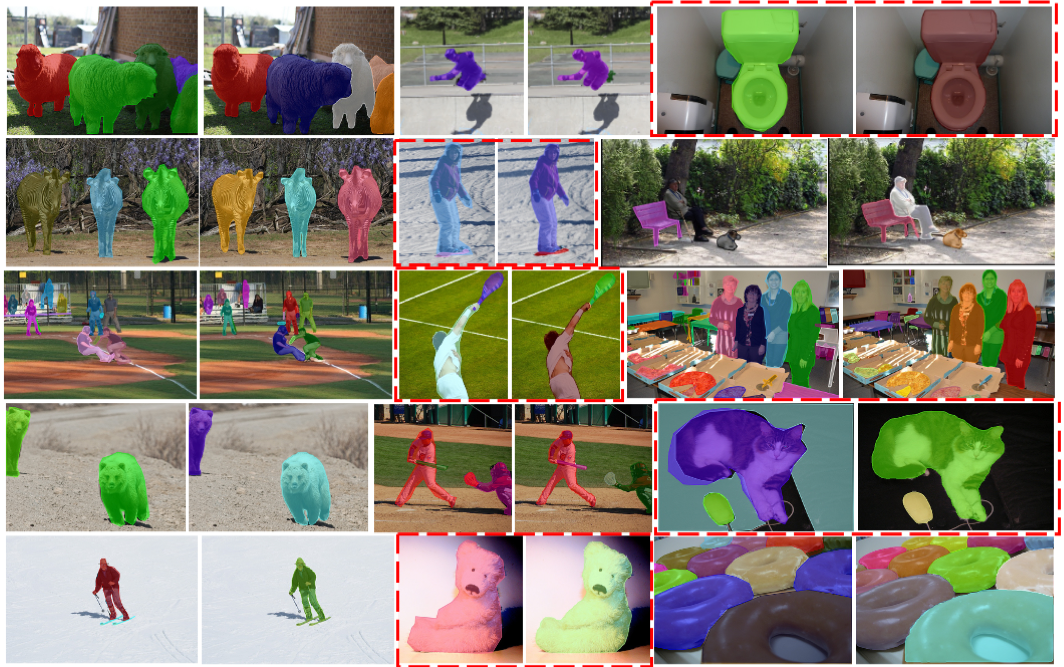}
	\caption{Visualization pairs of ground truth~(left image) and our method~(right image). Best viewed on the screen and zoom in.}
	\label{fig:coco_vis_supp}
\end{figure*}

Second, To show the smooth boundaries and the accurate segmentation results produced by our approach, some image pairs consist of the visualization result of our model and the corresponding ground truth on the COCO dataset~\cite{COCO} are provided in Fig.~\ref{fig:coco_vis_supp}. Note that in some situations, the boundaries and masks generated by our model are much better than the ground truth's boundaries, such as the third pair in the first row, the second pair in the second row, the second pair in the third row, the third pair in the fourth row, and the second pair in the fifth row.

\noindent \textbf{Results on Cityscapes and LVIS datasets: }Besides the COCO dataset, the Cityscape~\cite{Cityscapes} and LVIS~\cite{gupta2019lvis} datasets are also adopted to verify the effectiveness and generalization of BSM. The annotations of both datasets have significantly higher quality, especially on the boundary. In both datasets, our method is compared with the baseline model Mask R-CNN~\cite{MaskR-CNN} and two recent methods, including PointRend~\cite{PointRend} and BMask R-CNN~\cite{BMaskR-CNN}. Table.~\ref{tab:comparison_results_with_other_boundary_method_cityscape_lvis} shows the results on both datasets. Our BSM achieves the best results with ResNet-50 or ResNet-101 as the backbone network on both datasets.

\subsection{Experiment: Semantic Segmentation}

\begin{figure*}[h]
	\centering
	\includegraphics[width=1.0\linewidth]{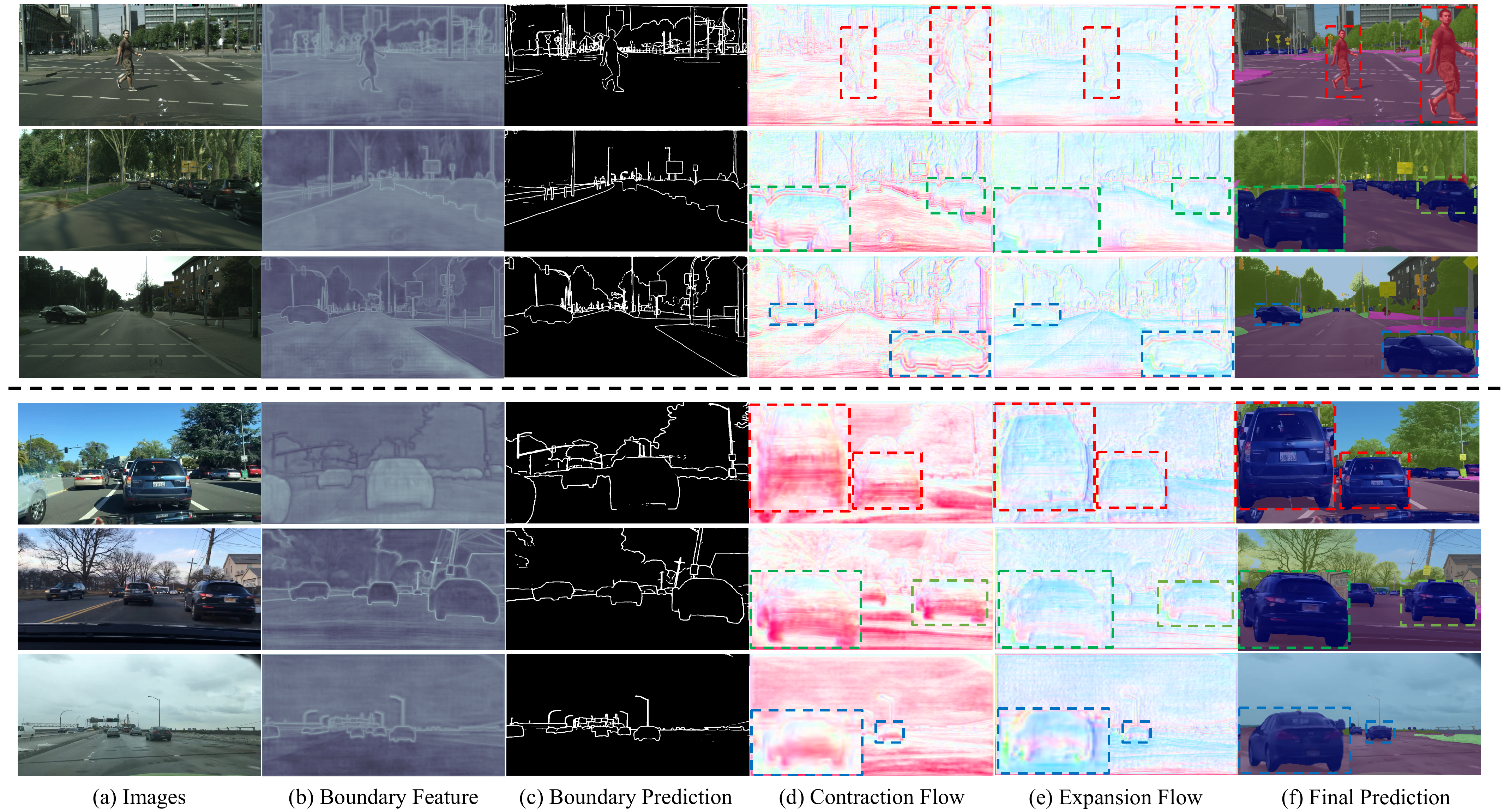}
	\caption{ Visualization of our BSM using DeeplabV3+ as baseline on Cityscapes. From left to right, (a) is the original image, we visualize the boundary feature in (b) and boundary prediction in (c), also two opposite flow field in (d) and (e) respectively. The final mask prediction is shown in (f). Best view it in color and zoom in. }
	\label{fig:semantic_vis}
\end{figure*}

\begin{table*}[!t]\setlength{\tabcolsep}{9pt}
	\centering
		\scalebox{1}{
			\begin{tabular}{l| c| c| c| c c c c }
			\hline
				Method   & backbone & dataset & mIoU $\uparrow$ & F1(12px) $\uparrow$ & F1(9px) $\uparrow$ & F1(5px) $\uparrow$ & F1(3px) $\uparrow$ \\
				\hline
		DeeplabV3+ \cite{deeplabv3p} &  R-50 & Cityscapes & 77.9 & 78.3 & 77.0 & 72.9 & 62.3 \\
    	+PointRend \cite{PointRend}  &  R-50 & Cityscapes & 78.5 & 79.6 & 78.5 & 74.8 & 64.1 \\
        +BSM   & R-50 &  Cityscapes & 79.5 & 80.9 & 79.7 & 76.4 & 67.7  \\
        \hline
        DeeplabV3+ \cite{deeplabv3p} &  R-101 & Cityscapes & 78.9 & 79.9 & 78.2 & 72.8 & 62.5\\
    	+PointRend \cite{PointRend}  & R-101 & Cityscapes & 79.5 & 80.7 & 78.6 & 74.8 & 64.6 \\
        +BSM   & R-101 &  Cityscapes  & 80.6 & 82.9 & 81.8 & 78.7 & 69.9  \\
        \hline
        DeeplabV3+ \cite{deeplabv3p} & R-50 & BDD & 61.2 & 76.6 & 75.4 & 70.8 & 61.6\\
    	+PointRend \cite{PointRend}  & R-50 & BDD & 62.3 & 79.1 & 77.3 & 72.6 & 63.9 \\
        +BSM   & R-50 &  BDD  & 63.5 & 81.1 & 79.9 & 76.2 & 67.9 \\
	\hline
	\end{tabular}}
		\caption{ Comparison results of DeeplabV3+, PointRend and BSM. All results are reported on the validation set. X-px means X pixels along the boundaries. R means ResNet. All the models are trained and tested in the same setting.}
		\label{tab:results_on_cityscapes_bdd}
\end{table*}

\noindent
\textbf{Overview} We also carry out experiments on semantic segmentation task to verify the generality of our BSM. Our BSM is inserted behind the ASPP module in DeeplabV3+~\cite{deeplabv3p}. 

\noindent \textbf{Dataset:} \textbf{(a) Cityscapes semantic segmentation set~\cite{Cityscapes}} , compared with Cityscapes instance segmentation set introduced in Sec.~\ref{sec:instance_segmentation_datasets_and_metrics}, Cityscapes semantic segmentation set has 19 categories. Both datasets have the same training, validation, and test images. \textbf{(b) BDD~\cite{BDD}} is a recently proposed road scene dataset with 10000 images in total. For dataset split, 7000, 1000, and 2000 images are used for train, validation and, test. For both datasets, our models are trained on their training set and report results on the validation set.

\noindent \textbf{Metrics: } In semantic segmentation, mean IoU~(mIoU) is the most commonly used metric. Besides, to evaluate the segmentation quality on the boundary, following~\cite{gatedSCNN,VOS_benchmark}, F-score is also reported on the boundary with four thresholds.

\noindent \textbf{Implementation details:} The DeeplabV3+~\cite{deeplabv3p} is employed as the baseline model. Following the original DeeplabV3+ paper, the output stride of backbone network is set to 8.  The same training and testing setting are used for different methods for a fair comparison. For both Cityscapes~\cite{Cityscapes} and BDD~\cite{BDD} datasets, stochastic gradient descent~(SGD) is adopted as the optimizer, and the batch size is set to 8~(one image per GPU) in both datasets. Momentum and weight decay are set to 0.9 and 0.0005, respectively. The initial learning rate is set to 0.01. Following the common practice, the `poly' learning rate policy is used to reduce the learning rate by  multiplying $(1-\frac{iteration}{total\_iterations})^{0.9}$. Random horizontal flip, random resize, and random crop are used as data augmentations. The scale range of random resize is [0.75, 2.0], and the crop size of random crop augmentation is 832. All the models are trained for 100 epochs. During testing, images are resized to 832 $\times$ 832. The architecture of BSM is the same as that for instance segmentation. This Deeplab-based model employs the res2 backbone feature as the low-level feature to provide detailed location information. 

\noindent
\textbf{Main Results}
Quantitatively, BSM is compared with DeeplabV3+ and PointRend in Table.~\ref{tab:results_on_cityscapes_bdd}. For the Cityscapes dataset, with ResNet50 and ResNet101 as the backbone network, our method obtains 1.0 and 1.1 mIoU gain over PointRend. The F-scores of our approach is better that DeeplabV3+ and PointRend too. Note that with the reduction of the F-score threshold, our method achieves a more significant performance gain, which means our model can generate more precise boundaries. For the BDD dataset, our BSM obtains consistent performance improvements as well, which proves the generalization of our approach.

\noindent
\textbf{Visualization Results} Some visualization results of our model on semantic segmentation task are demonstrated in Fig.~\ref{fig:semantic_vis}. The upper part of Fig.~\ref{fig:semantic_vis} is the visualization of the Cityscape dataset, while the lower part of Fig.~\ref{fig:semantic_vis} is the visualization of the BDD dataset. The first column of Fig.~\ref{fig:semantic_vis} is the original images. As shown in the second and third columns of Fig.~\ref{fig:semantic_vis}, we observe the clear and thin boundary feature and boundary prediction. The two opposite flow fields~(contraction flow field and expansion flow field) are shown in the fourth and fifth columns of Fig.~\ref{fig:semantic_vis}. The segmentation results of our method are shown in the last column of Fig.~\ref{fig:semantic_vis}. Besides, as shown in the right part of Fig.~\ref{fig:teaser_2}, compared with PointRend, our proposed BSM has better visual results on two different types of objects: things~(the first and the third rows) and scenes~(the second row). Both visual results prove our motivation: the boundary squeezing process can generate more accurate boundaries via learned flow field warping, then these precise boundaries contribute to generating better segmentation results.

	\section{Conclusions}
In this paper, we have proposed a new idea to model the segmentation tasks as boundary squeezing process. Accordingly, an opposite flow-based module named Boundary Squeeze module (BSM) has been designed. The supervisions of BSM can be obtained via the dilation and erosion operators using the existing annotations. The boundary squeezing process can be achieved the via flow-based warping. After using BSM, the feature contains precise boundary information, and the mask quality is much better. BSM has been verified on two different segmentation tasks, including instance segmentation and semantic segmentation. Extensive results have proven that our BSM outperforms various previous works in multiple settings. On top of that, the proposed module can also be naturally extended into several low-level vision segmentation tasks where the boundaries are crucial. One shortcoming of BoundarySqueeze is the limited resolution problem for some large objects which will be our future work.

\section{Acknowledgement} 
This research was supported by the National Key Research and Development Program of China under Grant No. 2018AAA0100400, No.2020YFB2103402, and the National Natural Science Foundation of China under Grants 62071466, 62076242, and 61976208. Thanks to the SenseTime Research for computing resource. We also acknowledge the help of Kuiyuan Yang for helpful discussion.

	{\small
		\bibliographystyle{ieee_fullname}
		\bibliography{bsm_reference}

\begin{thebibliography}{10}\itemsep=-1pt

\bibitem{boundaries_network_fields}
Gedas Bertasius, Jianbo Shi, and Lorenzo Torresani.
\newblock Semantic segmentation with boundary neural fields.
\newblock In {\em Proceedings of the IEEE conference on computer vision and
  pattern recognition}, pages 3602--3610, 2016.

\bibitem{cnn_random_wark}
Gedas Bertasius, Lorenzo Torresani, Stella~X Yu, and Jianbo Shi.
\newblock Convolutional random walk networks for semantic image segmentation.
\newblock In {\em Proceedings of the IEEE Conference on Computer Vision and
  Pattern Recognition}, pages 858--866, 2017.

\bibitem{yolact}
Daniel Bolya, Chong Zhou, Fanyi Xiao, and Yong~Jae Lee.
\newblock Yolact: Real-time instance segmentation.
\newblock In {\em Proceedings of the IEEE/CVF International Conference on
  Computer Vision}, pages 9157--9166, 2019.

\bibitem{CascadeRCNN}
Zhaowei Cai and Nuno Vasconcelos.
\newblock Cascade r-cnn: Delving into high quality object detection.
\newblock In {\em Proceedings of the IEEE conference on computer vision and
  pattern recognition}, pages 6154--6162, 2018.

\bibitem{sipmask}
Jiale Cao, Rao~Muhammad Anwer, Hisham Cholakkal, Fahad~Shahbaz Khan, Yanwei
  Pang, and Ling Shao.
\newblock Sipmask: Spatial information preservation for fast image and video
  instance segmentation.
\newblock In {\em Proceedings of the European Conference on Computer Vision},
  pages 1--18, 2020.

\bibitem{blendmask}
Hao Chen, Kunyang Sun, Zhi Tian, Chunhua Shen, Yongming Huang, and Youliang
  Yan.
\newblock Blendmask: Top-down meets bottom-up for instance segmentation.
\newblock In {\em Proceedings of the IEEE/CVF conference on computer vision and
  pattern recognition}, pages 8573--8581, 2020.

\bibitem{htc}
Kai Chen, Jiangmiao Pang, Jiaqi Wang, Yu Xiong, Xiaoxiao Li, Shuyang Sun,
  Wansen Feng, Ziwei Liu, Jianping Shi, Wanli Ouyang, et~al.
\newblock Hybrid task cascade for instance segmentation.
\newblock In {\em Proceedings of the IEEE/CVF Conference on Computer Vision and
  Pattern Recognition}, pages 4974--4983, 2019.

\bibitem{Task_edge_detection}
Liang-Chieh Chen, Jonathan~T Barron, George Papandreou, Kevin Murphy, and
  Alan~L Yuille.
\newblock Semantic image segmentation with task-specific edge detection using
  cnns and a discriminatively trained domain transform.
\newblock In {\em Proceedings of the IEEE conference on computer vision and
  pattern recognition}, pages 4545--4554, 2016.

\bibitem{deeplabv2}
Liang-Chieh Chen, George Papandreou, Iasonas Kokkinos, Kevin Murphy, and Alan~L
  Yuille.
\newblock Deeplab: Semantic image segmentation with deep convolutional nets,
  atrous convolution, and fully connected crfs.
\newblock {\em IEEE transactions on pattern analysis and machine intelligence},
  40(4):834--848, 2017.

\bibitem{deeplabv3}
Liang-Chieh Chen, George Papandreou, Florian Schroff, and Hartwig Adam.
\newblock Rethinking atrous convolution for semantic image segmentation.
\newblock {\em arXiv preprint arXiv:1706.05587}, 2017.

\bibitem{deeplabv3p}
Liang-Chieh Chen, Yukun Zhu, George Papandreou, Florian Schroff, and Hartwig
  Adam.
\newblock Encoder-decoder with atrous separable convolution for semantic image
  segmentation.
\newblock In {\em Proceedings of the European conference on computer vision},
  pages 801--818, 2018.

\bibitem{tensormask}
Xinlei Chen, Ross Girshick, Kaiming He, and Piotr Doll{\'a}r.
\newblock Tensormask: A foundation for dense object segmentation.
\newblock In {\em Proceedings of the IEEE/CVF International Conference on
  Computer Vision}, pages 2061--2069, 2019.

\bibitem{Supervised_edge_net}
Xier Chen, Yanchao Lian, Licheng Jiao, Haoran Wang, YanJie Gao, and Shi
  Lingling.
\newblock Supervised edge attention network for accurate image instance
  segmentation.
\newblock In {\em Proceedings of the European Conference on Computer Vision},
  pages 617--631, 2020.

\bibitem{boundaryIoU}
Bowen Cheng, Ross Girshick, Piotr Doll{\'a}r, Alexander~C Berg, and Alexander
  Kirillov.
\newblock Boundary iou: Improving object-centric image segmentation evaluation.
\newblock In {\em Proceedings of the IEEE/CVF Conference on Computer Vision and
  Pattern Recognition}, pages 15334--15342, 2021.

\bibitem{BMaskR-CNN}
Tianheng Cheng, Xinggang Wang, Lichao Huang, and Wenyu Liu.
\newblock Boundary-preserving mask r-cnn.
\newblock In {\em Proceedings of the European Conference on Computer Vision},
  pages 660--676, 2020.

\bibitem{Cityscapes}
Marius Cordts, Mohamed Omran, Sebastian Ramos, Timo Rehfeld, Markus Enzweiler,
  Rodrigo Benenson, Uwe Franke, Stefan Roth, and Bernt Schiele.
\newblock The cityscapes dataset for semantic urban scene understanding.
\newblock In {\em Proceedings of the IEEE conference on computer vision and
  pattern recognition}, pages 3213--3223, 2016.

\bibitem{dai2016instance}
Jifeng Dai, Kaiming He, Yi Li, Shaoqing Ren, and Jian Sun.
\newblock Instance-sensitive fully convolutional networks.
\newblock In {\em Proceedings of the European Conference on Computer Vision},
  pages 534--549, 2016.

\bibitem{dcn}
Jifeng Dai, Haozhi Qi, Yuwen Xiong, Yi Li, Guodong Zhang, Han Hu, and Yichen
  Wei.
\newblock Deformable convolutional networks.
\newblock In {\em Proceedings of the IEEE international conference on computer
  vision}, pages 764--773, 2017.

\bibitem{de2017semantic}
Bert De~Brabandere, Davy Neven, and Luc Van~Gool.
\newblock Semantic instance segmentation with a discriminative loss function.
\newblock {\em arXiv preprint arXiv:1708.02551}, 2017.

\bibitem{deng2018learning}
Ruoxi Deng, Chunhua Shen, Shengjun Liu, Huibing Wang, and Xinru Liu.
\newblock Learning to predict crisp boundaries.
\newblock In {\em Proceedings of the European Conference on Computer Vision},
  pages 562--578, 2018.

\bibitem{FlowNet}
Alexey Dosovitskiy, Philipp Fischer, Eddy Ilg, Philip Hausser, Caner Hazirbas,
  Vladimir Golkov, Patrick Van Der~Smagt, Daniel Cremers, and Thomas Brox.
\newblock Flownet: Learning optical flow with convolutional networks.
\newblock In {\em Proceedings of the IEEE international conference on computer
  vision}, pages 2758--2766, 2015.

\bibitem{evans2006morphological}
Adrian~N Evans and Xin~U Liu.
\newblock A morphological gradient approach to color edge detection.
\newblock {\em IEEE Transactions on Image Processing}, 15(6):1454--1463, 2006.

\bibitem{DANet}
Jun Fu, Jing Liu, Haijie Tian, Yong Li, Yongjun Bao, Zhiwei Fang, and Hanqing
  Lu.
\newblock Dual attention network for scene segmentation.
\newblock In {\em Proceedings of the IEEE/CVF Conference on Computer Vision and
  Pattern Recognition}, pages 3146--3154, 2019.

\bibitem{PGN_net}
Ke Gong, Xiaodan Liang, Yicheng Li, Yimin Chen, Ming Yang, and Liang Lin.
\newblock Instance-level human parsing via part grouping network.
\newblock In {\em Proceedings of the European Conference on Computer Vision},
  pages 770--785, 2018.

\bibitem{gupta2019lvis}
Agrim Gupta, Piotr Dollar, and Ross Girshick.
\newblock Lvis: A dataset for large vocabulary instance segmentation.
\newblock In {\em Proceedings of the IEEE/CVF Conference on Computer Vision and
  Pattern Recognition}, pages 5356--5364, 2019.

\bibitem{hariharan2014simultaneous}
Bharath Hariharan, Pablo Arbel{\'a}ez, Ross Girshick, and Jitendra Malik.
\newblock Simultaneous detection and segmentation.
\newblock In {\em Proceedings of the European Conference on Computer Vision},
  pages 297--312, 2014.

\bibitem{he2021enhanced}
Hao He, Xiangtai Li, Guangliang Cheng, Jianping Shi, Yunhai Tong, Gaofeng Meng,
  V{\'e}ronique Prinet, and Lubin Weng.
\newblock Enhanced boundary learning for glass-like object segmentation.
\newblock {\em arXiv preprint arXiv:2103.15734}, 2021.

\bibitem{MaskR-CNN}
Kaiming He, Georgia Gkioxari, Piotr Doll{\'a}r, and Ross Girshick.
\newblock Mask r-cnn.
\newblock In {\em Proceedings of the IEEE international conference on computer
  vision}, pages 2961--2969, 2017.

\bibitem{he2015delving}
Kaiming He, Xiangyu Zhang, Shaoqing Ren, and Jian Sun.
\newblock Delving deep into rectifiers: Surpassing human-level performance on
  imagenet classification.
\newblock In {\em Proceedings of the IEEE international conference on computer
  vision}, pages 1026--1034, 2015.

\bibitem{ResNet}
Kaiming He, Xiangyu Zhang, Shaoqing Ren, and Jian Sun.
\newblock Deep residual learning for image recognition.
\newblock In {\em Proceedings of the IEEE conference on computer vision and
  pattern recognition}, pages 770--778, 2016.

\bibitem{MSRCNN}
Zhaojin Huang, Lichao Huang, Yongchao Gong, Chang Huang, and Xinggang Wang.
\newblock Mask scoring r-cnn.
\newblock In {\em Proceedings of the IEEE/CVF Conference on Computer Vision and
  Pattern Recognition}, pages 6409--6418, 2019.

\bibitem{CCNet}
Zilong Huang, Xinggang Wang, Lichao Huang, Chang Huang, Yunchao Wei, and Wenyu
  Liu.
\newblock Ccnet: Criss-cross attention for semantic segmentation.
\newblock In {\em Proceedings of the IEEE/CVF International Conference on
  Computer Vision}, pages 603--612, 2019.

\bibitem{STN}
Max Jaderberg, Karen Simonyan, Andrew Zisserman, et~al.
\newblock Spatial transformer networks.
\newblock {\em Advances in neural information processing systems},
  28:2017--2025, 2015.

\bibitem{snakes}
Michael Kass, Andrew Witkin, and Demetri Terzopoulos.
\newblock Snakes: Active contour models.
\newblock {\em International journal of computer vision}, 1(4):321--331, 1988.

\bibitem{ke2021bcnet}
Lei Ke, Yu-Wing Tai, and Chi-Keung Tang.
\newblock Deep occlusion-aware instance segmentation with overlapping bilayers.
\newblock In {\em Proceedings of the IEEE/CVF Conference on Computer Vision and
  Pattern Recognition}, pages 4019--4028, 2021.

\bibitem{aaf}
Tsung-Wei Ke, Jyh-Jing Hwang, Ziwei Liu, and Stella~X Yu.
\newblock Adaptive affinity fields for semantic segmentation.
\newblock In {\em Proceedings of the European Conference on Computer Vision},
  pages 587--602, 2018.

\bibitem{kim2021devil}
Myungchul Kim, Sanghyun Woo, Dahun Kim, and In~So Kweon.
\newblock The devil is in the boundary: Exploiting boundary representation for
  basis-based instance segmentation.
\newblock In {\em Proceedings of the IEEE/CVF Winter Conference on Applications
  of Computer Vision}, pages 929--938, 2021.

\bibitem{instancecut}
Alexander Kirillov, Evgeny Levinkov, Bjoern Andres, Bogdan Savchynskyy, and
  Carsten Rother.
\newblock Instancecut: from edges to instances with multicut.
\newblock In {\em Proceedings of the IEEE Conference on Computer Vision and
  Pattern Recognition}, pages 5008--5017, 2017.

\bibitem{PointRend}
Alexander Kirillov, Yuxin Wu, Kaiming He, and Ross Girshick.
\newblock Pointrend: Image segmentation as rendering.
\newblock In {\em Proceedings of the IEEE/CVF conference on computer vision and
  pattern recognition}, pages 9799--9808, 2020.

\bibitem{kokkinos2015pushing}
Iasonas Kokkinos.
\newblock Pushing the boundaries of boundary detection using deep learning.
\newblock {\em arXiv preprint arXiv:1511.07386}, 2015.

\bibitem{decouple}
Xiangtai Li, Xia Li, Li Zhang, Guangliang Cheng, Jianping Shi, Zhouchen Lin,
  Shaohua Tan, and Yunhai Tong.
\newblock Improving semantic segmentation via decoupled body and edge
  supervision.
\newblock In {\em Proceedings of the European Conference on Computer Vision},
  pages 435--452, 2020.

\bibitem{Li2019GlobalAT}
Xiangtai Li, Li Zhang, Ansheng You, Maoke Yang, Kuiyuan Yang, and Yunhai Tong.
\newblock Global aggregation then local distribution in fully convolutional
  networks.
\newblock In {\em British Machine Vision Conference}, 2019.

\bibitem{EMANet}
Xia Li, Zhisheng Zhong, Jianlong Wu, Yibo Yang, Zhouchen Lin, and Hong Liu.
\newblock Expectation-maximization attention networks for semantic
  segmentation.
\newblock In {\em Proceedings of the IEEE/CVF International Conference on
  Computer Vision}, pages 9167--9176, 2019.

\bibitem{SGR_gcn}
Xiaodan Liang, Zhiting Hu, Hao Zhang, Liang Lin, and Eric~P Xing.
\newblock Symbolic graph reasoning meets convolutions.
\newblock {\em Advances in Neural Information Processing Systems},
  31:1853--1863, 2018.

\bibitem{fpn}
Tsung-Yi Lin, Piotr Doll{\'a}r, Ross Girshick, Kaiming He, Bharath Hariharan,
  and Serge Belongie.
\newblock Feature pyramid networks for object detection.
\newblock In {\em Proceedings of the IEEE conference on computer vision and
  pattern recognition}, pages 2117--2125, 2017.

\bibitem{focalloss}
Tsung-Yi Lin, Priya Goyal, Ross Girshick, Kaiming He, and Piotr Doll{\'a}r.
\newblock Focal loss for dense object detection.
\newblock In {\em Proceedings of the IEEE international conference on computer
  vision}, pages 2980--2988, 2017.

\bibitem{COCO}
Tsung-Yi Lin, Michael Maire, Serge Belongie, James Hays, Pietro Perona, Deva
  Ramanan, Piotr Doll{\'a}r, and C~Lawrence Zitnick.
\newblock Microsoft coco: Common objects in context.
\newblock In {\em Proceedings of the European Conference on Computer Vision},
  pages 740--755, 2014.

\bibitem{PAN}
Shu Liu, Lu Qi, Haifang Qin, Jianping Shi, and Jiaya Jia.
\newblock Path aggregation network for instance segmentation.
\newblock In {\em Proceedings of the IEEE conference on computer vision and
  pattern recognition}, pages 8759--8768, 2018.

\bibitem{liu2018affinity}
Yiding Liu, Siyu Yang, Bin Li, Wengang Zhou, Jizheng Xu, Houqiang Li, and Yan
  Lu.
\newblock Affinity derivation and graph merge for instance segmentation.
\newblock In {\em Proceedings of the European Conference on Computer Vision},
  pages 686--703, 2018.

\bibitem{fcn}
Jonathan Long, Evan Shelhamer, and Trevor Darrell.
\newblock Fully convolutional networks for semantic segmentation.
\newblock In {\em Proceedings of the IEEE conference on computer vision and
  pattern recognition}, pages 3431--3440, 2015.

\bibitem{maninis2017convolutional}
Kevis-Kokitsi Maninis, Jordi Pont-Tuset, Pablo Arbel{\'a}ez, and Luc Van~Gool.
\newblock Convolutional oriented boundaries: From image segmentation to
  high-level tasks.
\newblock {\em IEEE transactions on pattern analysis and machine intelligence},
  40(4):819--833, 2017.

\bibitem{DiceLoss}
Fausto Milletari, Nassir Navab, and Seyed-Ahmad Ahmadi.
\newblock V-net: Fully convolutional neural networks for volumetric medical
  image segmentation.
\newblock In {\em 2016 fourth international conference on 3D vision}, pages
  565--571. IEEE, 2016.

\bibitem{neven2019instance}
Davy Neven, Bert~De Brabandere, Marc Proesmans, and Luc~Van Gool.
\newblock Instance segmentation by jointly optimizing spatial embeddings and
  clustering bandwidth.
\newblock In {\em Proceedings of the IEEE/CVF Conference on Computer Vision and
  Pattern Recognition}, pages 8837--8845, 2019.

\bibitem{papari2011edge}
Giuseppe Papari and Nicolai Petkov.
\newblock Edge and line oriented contour detection: State of the art.
\newblock {\em Image and Vision Computing}, 29(2-3):79--103, 2011.

\bibitem{pytorch}
Adam Paszke, Sam Gross, Francisco Massa, Adam Lerer, James Bradbury, Gregory
  Chanan, Trevor Killeen, Zeming Lin, Natalia Gimelshein, Luca Antiga, et~al.
\newblock Pytorch: An imperative style, high-performance deep learning library.
\newblock {\em Advances in neural information processing systems},
  32:8026--8037, 2019.

\bibitem{deepsnake}
Sida Peng, Wen Jiang, Huaijin Pi, Xiuli Li, Hujun Bao, and Xiaowei Zhou.
\newblock Deep snake for real-time instance segmentation.
\newblock In {\em Proceedings of the IEEE/CVF Conference on Computer Vision and
  Pattern Recognition}, pages 8533--8542, 2020.

\bibitem{VOS_benchmark}
Federico Perazzi, Jordi Pont-Tuset, Brian McWilliams, Luc Van~Gool, Markus
  Gross, and Alexander Sorkine-Hornung.
\newblock A benchmark dataset and evaluation methodology for video object
  segmentation.
\newblock In {\em Proceedings of the IEEE conference on computer vision and
  pattern recognition}, pages 724--732, 2016.

\bibitem{fasterRCNN}
Shaoqing Ren, Kaiming He, Ross Girshick, and Jian Sun.
\newblock Faster r-cnn: Towards real-time object detection with region proposal
  networks.
\newblock {\em Advances in neural information processing systems}, 28:91--99,
  2015.

\bibitem{imagenet}
Olga Russakovsky, Jia Deng, Hao Su, Jonathan Krause, Sanjeev Satheesh, Sean Ma,
  Zhiheng Huang, Andrej Karpathy, Aditya Khosla, Michael Bernstein, et~al.
\newblock Imagenet large scale visual recognition challenge.
\newblock {\em International journal of computer vision}, 115(3):211--252,
  2015.

\bibitem{deepcontour}
Wei Shen, Xinggang Wang, Yan Wang, Xiang Bai, and Zhijiang Zhang.
\newblock Deepcontour: A deep convolutional feature learned by positive-sharing
  loss for contour detection.
\newblock In {\em Proceedings of the IEEE conference on computer vision and
  pattern recognition}, pages 3982--3991, 2015.

\bibitem{song2018cumulative}
Jingkuan Song, Zhilong Zhou, Lianli Gao, Xing Xu, and Heng~Tao Shen.
\newblock Cumulative nets for edge detection.
\newblock In {\em Proceedings of the 26th ACM international conference on
  Multimedia}, pages 1847--1855, 2018.

\bibitem{gatedSCNN}
Towaki Takikawa, David Acuna, Varun Jampani, and Sanja Fidler.
\newblock Gated-scnn: Gated shape cnns for semantic segmentation.
\newblock In {\em Proceedings of the IEEE/CVF International Conference on
  Computer Vision}, pages 5229--5238, 2019.

\bibitem{CondInst}
Zhi Tian, Chunhua Shen, and Hao Chen.
\newblock Conditional convolutions for instance segmentation.
\newblock In {\em Proceedings of the European Conference on Computer Vision},
  pages 282--298, 2020.

\bibitem{scnet}
Thang Vu, Haeyong Kang, and Chang~D Yoo.
\newblock Scnet: Training inference sample consistency for instance
  segmentation.
\newblock In {\em Proceedings of the AAAI Conference on Artificial
  Intelligence}, volume~35, pages 2701--2709, 2021.

\bibitem{nonlocal}
Xiaolong Wang, Ross Girshick, Abhinav Gupta, and Kaiming He.
\newblock Non-local neural networks.
\newblock In {\em Proceedings of the IEEE conference on computer vision and
  pattern recognition}, pages 7794--7803, 2018.

\bibitem{solo}
Xinlong Wang, Tao Kong, Chunhua Shen, Yuning Jiang, and Lei Li.
\newblock Solo: Segmenting objects by locations.
\newblock In {\em Proceedings of the European Conference on Computer Vision},
  pages 649--665, 2020.

\bibitem{solov2}
Xinlong Wang, Rufeng Zhang, Tao Kong, Lei Li, and Chunhua Shen.
\newblock Solov2: Dynamic and fast instance segmentation.
\newblock {\em Advances in Neural Information Processing Systems}, 33, 2020.

\bibitem{wold1987principal}
Svante Wold, Kim Esbensen, and Paul Geladi.
\newblock Principal component analysis.
\newblock {\em Chemometrics and intelligent laboratory systems}, 2(1-3):37--52,
  1987.

\bibitem{forestrcnn}
Jialian Wu, Liangchen Song, Tiancai Wang, Qian Zhang, and Junsong Yuan.
\newblock Forest r-cnn: Large-vocabulary long-tailed object detection and
  instance segmentation.
\newblock In {\em Proceedings of the 28th ACM International Conference on
  Multimedia}, pages 1570--1578, 2020.

\bibitem{detectron2}
Yuxin Wu, Alexander Kirillov, Francisco Massa, Wan-Yen Lo, and Ross Girshick.
\newblock Detectron2.
\newblock \url{https://github.com/facebookresearch/detectron2}, 2019.

\bibitem{upernet}
Tete Xiao, Yingcheng Liu, Bolei Zhou, Yuning Jiang, and Jian Sun.
\newblock Unified perceptual parsing for scene understanding.
\newblock In {\em Proceedings of the European Conference on Computer Vision},
  pages 418--434, 2018.

\bibitem{polarmask}
Enze Xie, Peize Sun, Xiaoge Song, Wenhai Wang, Xuebo Liu, Ding Liang, Chunhua
  Shen, and Ping Luo.
\newblock Polarmask: Single shot instance segmentation with polar
  representation.
\newblock In {\em Proceedings of the IEEE/CVF conference on computer vision and
  pattern recognition}, pages 12193--12202, 2020.

\bibitem{xie2017aggregated}
Saining Xie, Ross Girshick, Piotr Doll{\'a}r, Zhuowen Tu, and Kaiming He.
\newblock Aggregated residual transformations for deep neural networks.
\newblock In {\em Proceedings of the IEEE conference on computer vision and
  pattern recognition}, pages 1492--1500, 2017.

\bibitem{holistically}
Saining Xie and Zhuowen Tu.
\newblock Holistically-nested edge detection.
\newblock In {\em Proceedings of the IEEE international conference on computer
  vision}, pages 1395--1403, 2015.

\bibitem{BDD}
Fisher Yu, Haofeng Chen, Xin Wang, Wenqi Xian, Yingying Chen, Fangchen Liu,
  Vashisht Madhavan, and Trevor Darrell.
\newblock Bdd100k: A diverse driving dataset for heterogeneous multitask
  learning.
\newblock In {\em Proceedings of the IEEE/CVF conference on computer vision and
  pattern recognition}, pages 2636--2645, 2020.

\bibitem{dilation}
Fisher Yu and Vladlen Koltun.
\newblock Multi-scale context aggregation by dilated convolutions.
\newblock In {\em 4th International Conference on Learning Representations},
  2016.

\bibitem{casenet}
Zhiding Yu, Chen Feng, Ming-Yu Liu, and Srikumar Ramalingam.
\newblock Casenet: Deep category-aware semantic edge detection.
\newblock In {\em Proceedings of the IEEE conference on computer vision and
  pattern recognition}, pages 5964--5973, 2017.

\bibitem{ocrnet}
Yuhui Yuan, Xilin Chen, and Jingdong Wang.
\newblock Object-contextual representations for semantic segmentation.
\newblock In {\em Proceedings of the European Conference on Computer Vision},
  pages 173--190, 2020.

\bibitem{dgmn}
Li Zhang, Dan Xu, Anurag Arnab, and Philip~HS Torr.
\newblock Dynamic graph message passing networks.
\newblock In {\em Proceedings of the IEEE/CVF Conference on Computer Vision and
  Pattern Recognition}, pages 3726--3735, 2020.

\bibitem{MEInst}
Rufeng Zhang, Zhi Tian, Chunhua Shen, Mingyu You, and Youliang Yan.
\newblock Mask encoding for single shot instance segmentation.
\newblock In {\em Proceedings of the IEEE/CVF Conference on Computer Vision and
  Pattern Recognition}, pages 10226--10235, 2020.

\bibitem{PSPNet}
Hengshuang Zhao, Jianping Shi, Xiaojuan Qi, Xiaogang Wang, and Jiaya Jia.
\newblock Pyramid scene parsing network.
\newblock In {\em Proceedings of the IEEE conference on computer vision and
  pattern recognition}, pages 2881--2890, 2017.

\bibitem{ADE20K}
Bolei Zhou, Hang Zhao, Xavier Puig, Tete Xiao, Sanja Fidler, Adela Barriuso,
  and Antonio Torralba.
\newblock Semantic understanding of scenes through the ade20k dataset.
\newblock {\em International Journal of Computer Vision}, 127(3):302--321,
  2019.

\bibitem{dcnv2}
Xizhou Zhu, Han Hu, Stephen Lin, and Jifeng Dai.
\newblock Deformable convnets v2: More deformable, better results.
\newblock In {\em Proceedings of the IEEE/CVF Conference on Computer Vision and
  Pattern Recognition}, pages 9308--9316, 2019.

\bibitem{DFF}
Xizhou Zhu, Yuwen Xiong, Jifeng Dai, Lu Yuan, and Yichen Wei.
\newblock Deep feature flow for video recognition.
\newblock In {\em Proceedings of the IEEE conference on computer vision and
  pattern recognition}, pages 2349--2358, 2017.

\bibitem{video_propagation}
Yi Zhu, Karan Sapra, Fitsum~A Reda, Kevin~J Shih, Shawn Newsam, Andrew Tao, and
  Bryan Catanzaro.
\newblock Improving semantic segmentation via video propagation and label
  relaxation.
\newblock In {\em Proceedings of the IEEE/CVF Conference on Computer Vision and
  Pattern Recognition}, pages 8856--8865, 2019.

\end{thebibliography}
	}
	
\end{document}